\pdfoutput=1
\documentclass[11pt]{article}

\usepackage[final]{acl}

\usepackage{fancyhdr}

\fancypagestyle{emnlpfirstpage}{%
  \fancyhf{}

  \fancyfoot[C]{%
    \parbox[t]{\textwidth}{%
      \centering
      {\fontsize{10}{12}\selectfont\thepage}\\[4pt]
      \fontsize{8}{10}\selectfont
      \textit{Proceedings of the 2026 Conference on Empirical Methods in Natural Language Processing}, pages 26339--26365\\
      October 24--29, 2026 \textcopyright{}2026 Association for Computational Linguistics
    }%
  }
}

\usepackage{times}
\usepackage{soul}
\usepackage{footmisc}

\usepackage{marvosym}
\usepackage{latexsym}
\usepackage{graphicx}
\graphicspath{{../}}
\usepackage{booktabs, multirow}
\usepackage{listings}
\usepackage[most]{tcolorbox}
\usepackage{algpseudocode}
\usepackage{amsmath} 
\usepackage{amsfonts}
\usepackage{xcolor}
\usepackage{tabularx}
\usepackage[framemethod=TikZ]{mdframed}
\usepackage{adjustbox}
\mdfsetup{
  roundcorner=10pt,
  linewidth=1pt,
  innertopmargin=10pt,
  innerbottommargin=10pt,
  innerrightmargin=10pt,
  innerleftmargin=10pt,
  backgroundcolor=gray!10
}
\usepackage{subcaption}
\usepackage{makecell}
\usepackage{enumitem}
\usepackage{amssymb}
\usepackage[
linesnumbered,
ruled,vlined,noend]{algorithm2e}

\usepackage[table,xcdraw]{xcolor}
\lstdefinelanguage{json}{
    basicstyle=\footnotesize\ttfamily,
    numberstyle=\scriptsize,
    stepnumber=1,
    numbersep=8pt,
    showstringspaces=false,
    breaklines=true,
    frame=none, % No frame around the code
    backgroundcolor=\color{white}, % Reset background color to white or transparent
    literate=
     *{:}{{{\color{purple}{:}}}}{1}
      {,}{{{\color{purple}{,}}}}{1}
      {\{}{{{\color{blue}{\{}}}}{1}
      {\}}{{{\color{blue}{\}}}}}{1}
      {[}{{{\color{blue}{[}}}}{1}
      {]}{{{\color{blue}{]}}}}{1}
      {β}{{{\ensuremath{\beta}}}}{1},
}
\definecolor{mygreen}{RGB}{0, 128, 0} % 定义一个深绿色
\definecolor{lightblue}{RGB}{92, 144, 198}
\newcommand{\gain}[1]{\textbf{\textcolor{mygreen}{#1}}}

\usepackage{amsthm}
\theoremstyle{definition}
\newtheorem{theorem}{Theorem}[section]
\newtheorem{definition}[theorem]{Definition}

\usepackage[T1]{fontenc}
\usepackage[utf8]{inputenc}

\usepackage{microtype}

\usepackage{inconsolata}
\usepackage{float}
\usepackage{CJKutf8}

\title{
Patient-Zero: Scaling Synthetic Patient Agents to\\Real-World Distributions without Real Patient Data
}

\author{
\textbf{Yunghwei Lai}$^{1,2}$, \textbf{Ziyue Wang}$^{1,2}$, \textbf{Weizhi Ma}$^{2,\dagger}$, \textbf{Yang Liu}$^{1,2,\dagger}$\\
  \textsuperscript{1}Dept. of Comp. Sci. \& Tech., Institute for AI, Tsinghua University, Beijing, China \\
  \textsuperscript{2}Institute for AI Industry Research (AIR), Tsinghua University, Beijing, China\\
  \texttt{\fontsize{11pt}{10pt}\selectfont {mawz}@tsinghua.edu.cn, {liuyang2011}@tsinghua.edu.cn}
  }

\DefineFNsymbols*{1}{*}
\setfnsymbol{1}

\begin{document}
\setcounter{page}{26339}          % 起始页码，与截图一致
\pagestyle{plain}               % 后续页面显示普通页码

\maketitle
\thispagestyle{emnlpfirstpage}   % 仅首页显示会议出版信息

% \renewcommand{\thefootnote}{\fnsymbol{footnote}} 
%     \footnotetext[1]{Equal contribution.}
% \renewcommand{\thefootnote}{\arabic{footnote}}

\noindent\let\thefootnote\relax\footnotetext{$^\dagger$ Corresponding Authors.}
\noindent\let\thefootnote\relax\footnotetext{$^\ddagger$ Code, data and model weights are available at \url{https://github.com/thu-unicorn/Patient-Zero}.}

% \DefineFNsymbols*{1}{\Letter}
% \setfnsymbol{1}
% \renewcommand{\thefootnote}{\fnsymbol{footnote}} 
%     \footnotetext[1]{Corresponding authors.}
% \renewcommand{\thefootnote}{\arabic{footnote}}

\begin{abstract}
Synthetic data generation with Large Language Models (LLMs) has emerged as a promising solution in the medical domain to mitigate data scarcity and privacy constraints. However, existing approaches remain constrained by their derivative nature, relying on real-world records, which pose privacy risks and distribution biases. Furthermore, current patient agents face the \textit{Stability-Plasticity Dilemma}, struggling to maintain clinical consistency during dynamic inquiries. To address these challenges, we introduce \textbf{Patient-Zero}, a novel framework for \textit{ab initio} patient simulation that requires \textit{no real medical records}. Our Medically-Aligned Hierarchical Synthesis framework generates comprehensive and diverse patient records from medical knowledge via stratified attribute permutation. To support rigorous clinical interaction, we design a Dual-Track Cognitive Memory System to enable agents dynamically update memory while preserving logical consistency and persona adherence. 
Extensive evaluations show that \textbf{Patient-Zero} outperforms baselines in both data quality and interaction fidelity. 
In physician evaluations, our records obtain the highest mean diversity, consistency, and completeness ratings. Downstream medical reasoning model trained on our synthetic dataset shows substantial performance gains (MedQA +24.0\%; MMLU +14.5\%), demonstrating the practical utility of our framework.

% Evaluations show that Patient-Zero improves synthetic record quality and interaction fidelity under our defined clinical narrative and dialogue metrics. In blind physician evaluations, our records were frequently judged as human-authored and received higher ratings on structural completeness, while we explicitly distinguish this from fully reproducing the noise and heterogeneity of real clinical populations. Furthermore, training on Patient-Zero data yields downstream gains on medical reasoning benchmarks, suggesting practical utility as a privacy-free training resource.
\end{abstract}

\begin{figure}[h!]
    \centering
    \includegraphics[width=\linewidth]{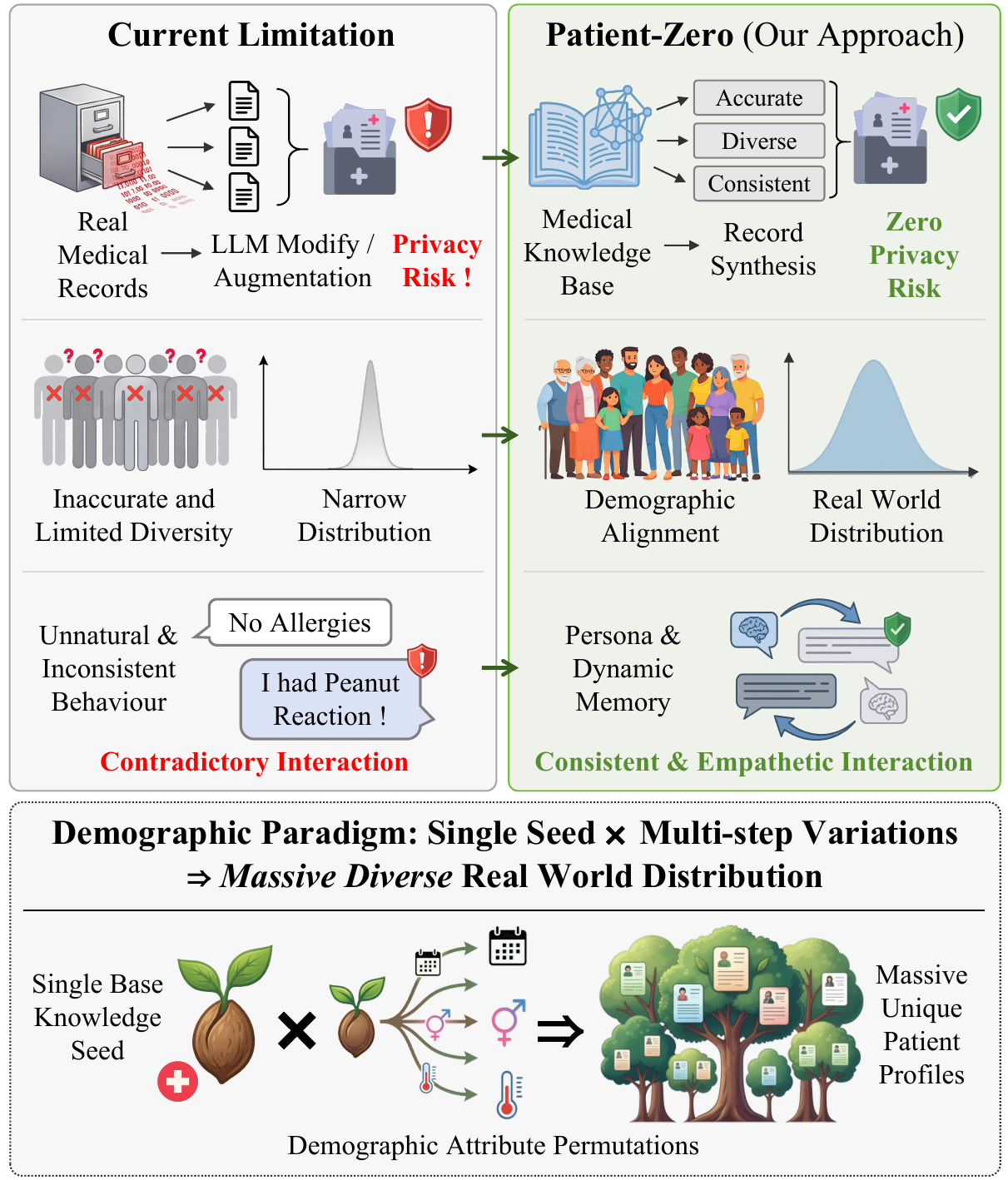}
    \vspace{-21pt}
    \caption{Our \textbf{Patient-Zero} Paradigm. While conventional methods are constrained by the derivative nature of real-world data, such as privacy risks and distribution biases, our framework enables \textit{ab initio} patient simulation. Instead of using sensitive medical records as seed, \textbf{Patient-Zero} constructs patient agents \textit{from scratch} using medical knowledge, achieving zero privacy risk while maintaining clinical consistency throughout synthetic data generation and interactive simulation.}
    \label{fig:main}
    % \vspace{-18pt}
\end{figure}

\vspace{-10pt}
\section{Introduction}
% Background on synthetic patient record generation.
% Importance of multi-step generation in medical data simulation.
% Objectives: Improving diagnostic accuracy, data quality, and alignment with medical theory.

% \begin{figure*}[htbp]
%     \centering
%     \includegraphics[width=\textwidth]{figure/flowchart.pdf}
%     \caption{The proposed framework consists of two main modules: the Patient Record Construction module and the Patient Agent Interaction Simulation module. Patient records are generated through a multi-step process with atomic fact decomposition, using a knowledge base and stored in patient agent memory. In the Patient Agent Interaction Simulation module, these patient records are utilized to simulate dialogues between patient agents and doctors. Patient responses are evaluated using a triplet evaluation mechanism, where consistent responses are replied directly, neutral statements dynamically update the patient agent memory, and contradictory responses trigger regeneration.}
%     \label{fig:flowchart}
% \end{figure*}

% 介绍背景
Large Language Models (LLMs) have shown remarkable capabilities in generative tasks~\cite{openai2024gpt4technicalreport, zhang2025survey}, increasingly being used for synthetic data construction across various domains~\cite{li-etal-2023-synthetic, guo2024generativeaisyntheticdata, nahid2024safesynthdpleveraginglargelanguage, karst2024generativeaibanksbenchmarks}. In the medical field, high-quality synthetic data offers a promising solution for high costs and strict privacy constraints in real-world patient data~\cite{Goncalves2020}.
Dominant approaches for patient record generation span retrieval-reasoning~\cite{xu2025retrievalreasoninglargelanguagemodelbased}, reinforcement learning (RL)~\cite{das2024synrlaligningsyntheticclinical}, generative adversarial network (GAN)~\cite{LDP-GAN, Li2023EHRMGAN}, and alternative methods~\cite{sun2023collaborativesynthesispatientrecords, Avatar, tornqvist2024texttotabularapproachgeneratesynthetic, liu2024hcllmhistoricalconstrainedlargelanguage, synthea}.

% 现有缺陷
However, existing approaches are fundamentally constrained by two primary limitations. First, regarding \textbf{data quality}, methods derived from \textit{real-world medical records} face inherent bottlenecks: \textbf{privacy risks}, \textbf{medical inconsistency}, and \textbf{limited diversity}. Even with de-identification methods to exclude personal identifiers, generative approaches may still pose privacy risks~\cite{Chen2024GenerativeAIPrivacy} or fail to strictly adhere to medical knowledge~\cite{Chen2025HallucinationsAI}. Furthermore, the reliance on specific source datasets restricts the distribution, failing to capture the full complexity of real-world scenarios~\cite{Giuffre2023SyntheticHealthcare}. Second, in terms of \textbf{interactive capability}, existing patient agents often suffer from unnatural and inconsistent behaviors in clinical inquiries~\cite{Graf2024VirtualPatients, Cook2025VirtualPatientsLLM}, or are confined to specific domains~\cite{patientpsi, roleplay-doh, psyche}, rendering them unable to generalize across multi-specialty clinical scenarios. This motivates a fundamental research question:
% \vspace{-3pt}

\begin{center} \textbf{\textit{Can we generate realistic interactive patient agents without exposing any real-world records?}} \end{center}
In this work, we propose \textbf{Patient-Zero} as an effective solution to this challenge, which is a novel hierarchical generation framework that synthesizes medically-aligned patient records without relying on any real patient data. It employs a hierarchical process anchored in medical knowledge, leveraging \textit{epidemiological attribute permutations} to evolve abstract disease concepts 
% (base knowledge seeds) 
into granular patient records. To bridge the gap between static records and dynamic interaction, we introduce a Natural Language Inference Verifier (\textit{NLI-Verifier}), ensuring both logical consistency and persona adherence for robust memory updates. To validate the practical utility of this framework, manual evaluations by licensed physicians confirm that \textbf{Patient-Zero} produces narratives that strictly adhere to professional standards.
% Furthermore, we deploy \textbf{Patient-Zero} as the core patient simulation engine in \textit{Agent Hospital}~\cite{li2024agenthospitalsimulacrumhospital} for large-scale realistic interactions for medical agent training.
Our key contributions are summarized as follows:
% \vspace{-6pt}
% \begin{itemize}[left=0cm]
\begin{itemize}[left=0cm, itemsep=2pt, parsep=0pt]
    \item \textbf{(Framework)} We introduce \textbf{Patient-Zero}, a novel framework that generates large-scale synthetic patient records \textit{ab initio}, without utilizing any real patient data, 
    % across multi-specialty domains. By integrating a guideline-anchored hierarchical process with \textit{epidemiological attribute permutations}, our approach ensures 
    ensuring strict clinical adherence while achieving scalable data diversity that aligns with real-world distributions.
    \item \textbf{(Simulation)}  We propose a \textit{Dual-Track Cognitive Memory System} to resolve interaction inconsistencies. Patient agents can perform memory updates with logical consistency and persona adherence during extended clinical dialogues.
    \item \textbf{(Validation)} 
    Extensive evaluations demonstrate that \textbf{Patient-Zero} substantially outperforms baselines in synthetic data quality and interaction fidelity, and aligning with epidemiological marginals. In blind licensed physician evaluations, our records obtain the highest descriptive overall score and the highest mean diversity, consistency, and completeness ratings.
    \item \textbf{(Utility)} \textbf{Patient-Zero} supports medical education and OSCE-style simulation through repeatable communication practice, while providing scalable, controllable scenarios for downstream model training. The results on medical-student user studies and medical benchmarks (MedQA, MMLU) support the utility of our framework.
    % We evaluate Patient-Zero across synthetic record quality, dynamic interaction fidelity, distributional alignment, expert physician review, and downstream model training. The results show that our synthetic records are clinically coherent under expert-defined criteria, and aligned with selected epidemiological marginals and clinically motivated joint patterns.
\end{itemize}

\begin{figure*}
    \centering
    \includegraphics[width=\textwidth]{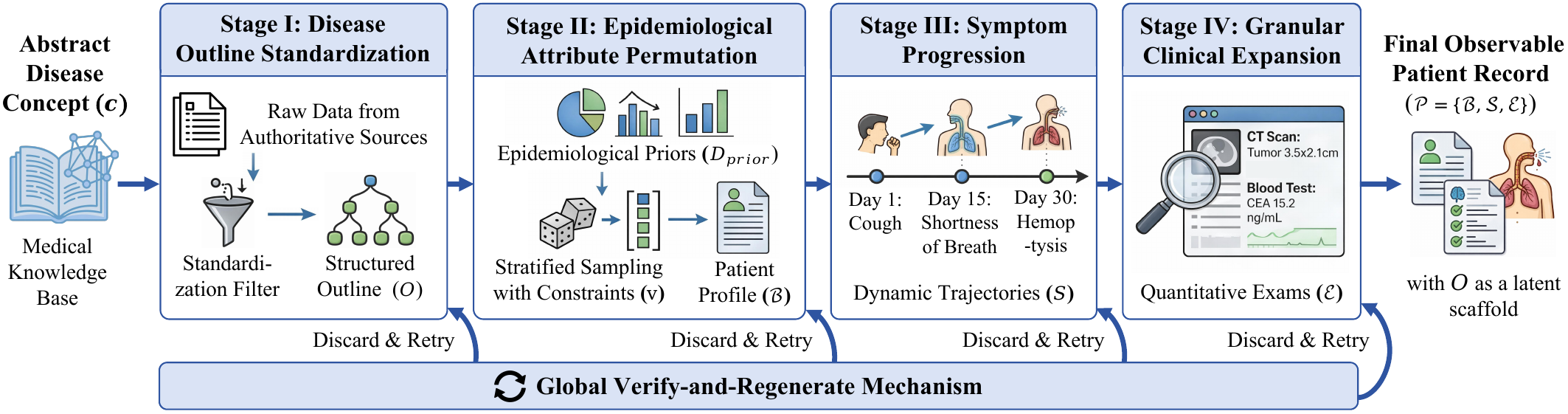}
    \vspace{-18pt}
    \caption{Overview of our \textbf{Patient-Zero} Hierarchical Synthesis Framework. The pipeline factorizes patient generation into a four-stage causal chain ($c \to \mathcal{O} \to \mathcal{B} \to \mathcal{S} \to \mathcal{E}$). Starting from an abstract disease concept ($c$), the system progressively expands medical details: I) Standardizing noisy knowledge into a structured outline; II) Sampling epidemiological attributes via constrained permutation; III) Evolving dynamic symptom trajectories; and IV) Generating granular, quantitative examination results. A global \textit{verify-and-regenerate} mechanism (bottom bar) enforces strict validity via iterative self-correction at every stage to prevent error propagation down the causal chain.}
    \label{fig:method}
    % \vspace{-12pt}
\end{figure*}

% \vspace{-1em}
\section{Related Work}
\label{sec:related_work}
% \vspace{-0.2em}
\paragraph{Synthetic Medical Record Generation}
In the healthcare sector, strict privacy regulations have significantly restricted access to real Electronic Health Records (EHRs), driving the research into synthetic data generation~\cite{Kruse_Smith_Vanderlinden_Nealand_2017, Goncalves2020-zp}. Early generative approaches predominantly utilized GAN~\cite{medgan, feng2024enhancingmedicalimaginggans, health-sensor}. With the paradigm shift toward LLM, models are conditioned on seed data retrieved from real-world records, including methods such as retrieval-reasoning~\cite{xu2025retrievalreasoninglargelanguagemodelbased}, RL~\cite{das2024synrlaligningsyntheticclinical}, and other approaches~\cite{sun2023collaborativesynthesispatientrecords, Yu2025SimulatedPatients, liu2024hcllmhistoricalconstrainedlargelanguage, Avatar, tornqvist2024texttotabularapproachgeneratesynthetic, synthea}. However, they remain fundamentally dependent on real-world patient data, which poses privacy risks and limits the diversity to the distribution of the source datasets. In contrast, \textbf{Patient-Zero} generates patient records \textit{without real private data} while \textit{aligning with diverse real-world distribution}.

\paragraph{Patient Agent Simulation}
Simulated patient agents serve as a scalable environment for training medical professionals and evaluating diagnostic reasoning~\cite{lizée2024conversationalmedicalaiready, zhu2024potentialllmsmedicaleducation}. Current methodologies generally fall into two categories. The first prioritizes behavioral and psychological fidelity, such as cognitive models and role-playing scenarios~\cite{patientpsi, roleplay-doh, psyche, patient-agent-coevolution, script-dialog-policy}, while the second emphasizes clinical accuracy and reasoning reliability, ensuring the agent provides factually correct and logically sound responses~\cite{Yu2025SimulatedPatients, mediq}. Nevertheless, strictly scripted agents often appear robotic, whereas open-ended generative models tend to forget their initial clinical persona or contradict themselves during extended interactions. \textbf{Patient-Zero} address this by implementing a real-time memory verification mechanism, maintaining strict logical consistency and persona adherence while preserving natural dialogue flexibility.

\section{The Patient-Zero Framework}
\label{sec:method}
% \vspace{-0.2em}
We propose \textbf{Patient-Zero}, a framework designed to synthesize realistic patient agents without referencing real-world medical records. Formally, we define the framework through two coupled modules: a \textit{Medically-Aligned Hierarchical Synthesis} protocol (Section \ref{subsec:synthesis}) that constructs static patient records, and an \textit{Dual-Track Cognitive Memory System} (Section \ref{subsec:memory_architecture}) that controls the agent behavior and memory evolution via logic verification.

% \vspace{-0.2em}
\subsection{Medically-Aligned Hierarchical Synthesis}
\label{subsec:synthesis}

% 整体框架介绍
Traditional approaches formulate patient generation as a derivative task, using existing real-world patient records $p(x_{patient}|x_{record})$. In contrast, \textbf{Patient-Zero} models the generation as an \textit{ab initio} synthesis process grounded in abstract medical concepts. Let $\mathcal{K}$ denote a medical knowledge base. The synthesis of a patient record $\mathcal{P}$ is factorized into a chain of conditional dependencies:
% \vspace{-0.6em}
\begin{equation}
\begin{aligned}
    p(\mathcal{P} | c) = p(\mathcal{E} | \mathcal{S}, \mathcal{B}, \mathcal{O}) \cdot p(\mathcal{S} | \mathcal{B}, \mathcal{O}) \\ \cdot p(\mathcal{B} | \mathcal{O}) \cdot p(\mathcal{O} | c)
\end{aligned}
\end{equation}

% \vspace{-0.6em}
\noindent
where $c \in \mathcal{K}$ represents the abstract disease concept (base knowledge seed). The process unfolds in four stages, as illustrated in Figure \ref{fig:method}. 
Each stage uses a \textit{verify-and-regenerate} mechanism. Deterministic checks handle schemas, demographic and anatomical compatibility, temporal order, physiological ranges, and arithmetic relations. An LLM semantic verifier checks cross-section relations that are difficult to encode. Failed sections are regenerated locally within a fixed retry budget.

% Comparing with previous medical record augmentation or generation methods, \textbf{Patient-Zero} utilizes only disease related knowledge to generate medical records. Though no privacy information is needed, it is more challenging to construct a comprehensive patient records from scratch, where the patient's basic information, epidemiology, symptoms, and clinical examination results should be included. 
% To achieve accurate and diverse patient records generation, we firstly propose a medically-aligned multi-step generation strategy, which is shown in Figure~\ref{fig:main_generation_step}. Note that the disease knowledge base can be easily collected from various websites (e.g., Wikipedia\footnote{\url{https://www.wikipedia.org/}}, Baidu Health Encyclopedia\footnote{\url{https://jiankang.baidu.com/}}).

\paragraph{Stage I: Disease Outline Standardization ($c \rightarrow \mathcal{O}$)}
Given a disease concept $c$ and its associated unstructured data retrieved from medical knowledge sources
% \footnote{\label{fn:wiki_source}Data sources include verified medical literature, reports and encyclopedias (e.g., Wikipedia Medical Portal).}
, we utilize an LLM to reconstruct a standardized disease outline $\mathcal{O}$. This stage functions as a \textit{knowledge standardization filter} which parses noisy input data into a unified structured schema anchored in medical knowledge, serving as a reliable reference for subsequent generation. The LLM structures disease knowledge guided by explicit controls, including fixed schemas, required fields, ontology terms, and rules that check whether each attribute is clinically compatible with the disease. Source provenance and authority criteria are detailed in Appendix~\ref{app:source-provenance}.

% \begin{figure}[htbp]
%     \centering
%     \includegraphics[width=\linewidth]{figure/long_multi_generation.pdf}
%     \caption{Multi-step Framework for Synthetic Patient Record Generation: A systematic approach starts with generating a disease outline from a knowledge base, followed by disease selection, basic patient information, and detailed examination results to create a complete synthetic patient record.}
%     \label{fig:long-multi-generation}
% \end{figure}

\paragraph{Stage II: Epidemiological Attribute Permutation ($\mathcal{O} \rightarrow \mathcal{B}$)}
To ensure the synthesized population follows selected epidemiological priors encoded by the knowledge base, we employ \textit{Stratified Sampling with Constraints} to explicitly decouple attribute selection from textual generation.
Specifically, we organize the high-dimensional attribute space $\mathcal{A}$ into a hierarchical taxonomy comprising four core dimensions based on the Social Determinants of Health (SDOH)~\cite{who_sdoh}. Each dimension $\mathcal{A}_k$ contains a set of specific fine-grained variables $\{x_1, x_2, \dots, x_m\}$ designed to comprehensively capture  components:
% Specifically, we formalize the patient record as a vector $\mathbf{v}$ in a high-dimensional attribute space $\mathcal{A}$. We categorize these attributes into four dimensions to comprehensively capture the Social Determinants of Health (SDOH)~\cite{who_sdoh}:
\textbf{1) Biological and Demographic ($\mathcal{A}_{bio}$):} Age strata, biological sex, physiological status, and ethnicity.
\textbf{2) Socioeconomic ($\mathcal{A}_{soc}$):} Geographic setting, and socioeconomic status indicators.
% \textbf{3) Family and Social ($\mathcal{A}_{fam}$):} Marital status, household structure, and dependents support.
\textbf{3) Behavioral and Lifestyle ($\mathcal{A}_{beh}$):} Substance use, communication, dietary and activity patterns.

Formally, we construct the patient vector $\mathbf{v}$ by concatenating the attribute values sampled from each dimension: $\mathbf{v} = \mathbf{v}_{bio} \oplus \mathbf{v}_{soc} \oplus \mathbf{v}_{fam} \oplus \mathbf{v}_{beh}$, where each sub-vector $\mathbf{v}_{k}$ consists of sampled values for the specific variables in that category.
% (e.g., $\mathbf{v}_{bio} = [\text{Age: 65},\ \text{Sex: Male},\cdots]$). 
Let $\mathcal{D}_{prior}$ denote the set of marginal distributions for these sub-attributes derived from medical knowledge. The sampling is formulated as:
% \vspace{-3pt}
\begin{equation}
    \mathbf{v} \sim \prod_{k} P(\mathbf{v}_k | \mathcal{O}, \mathcal{D}_{prior}) \cdot \mathbb{I}(\textsc{Valid}(\mathbf{v}))
    % \vspace{-3pt}
\end{equation}

% \vspace{-6pt}
\noindent
where $P(\mathbf{v}_k | \mathcal{O}, \mathcal{D}_{prior})$ denotes the set of categorical distributions weighted by real-world prevalence, and $\mathbb{I}(\textsc{Valid}(\mathbf{v}))$ is an indicator function that performs rejection sampling to eliminate logical inconsistencies. Conditioned on this fine-grained vector $\mathbf{v}$, the textual patient profile $\mathcal{B}$ is generated. This stage ensures \textbf{Patient-Zero} covers the \textit{``long tail''} of diverse populations while adhering to medical knowledge. Attribute sampling uses predefined priors, conditional distributions, and deterministic rejection rules without LLM generation. Detailed attribute taxonomies are provided in Appendix \ref{app:attributes}.

% Figure~\ref{fig:long-multi-generation} presents a synthetic patient record exemplifying a gout. %severe pancreatitis case.

% \vspace{-0.2em}
\paragraph{Stage III: Symptom Progression ($\mathcal{B}, \mathcal{O} \rightarrow \mathcal{S}$)}
To align with the standard clinical workflow, this stage generates the patient-specific symptoms $\mathcal{S}$ conditioned on the patient profile $\mathcal{B}$ and disease constraints $\mathcal{O}$.
We model symptoms as \textit{dynamic trajectories} characterized by onset, duration, severity, and triggers. 
Symptom generation is conditioned on patient-specific attributes,
% (e.g., a patient with a smoking history in $\mathcal{B}$ is more likely to manifest chronic cough or shortness of breath). 
creating a logical causal link between the patient's background and their clinical presentation.
 The LLM generates the narrative, while predefined disease stages, cause-effect links, and timing rules ensure that the patient trajectory remains clinically plausible.

% \vspace{-0.2em}
\paragraph{Stage IV: Granular Clinical Expansion ($\mathcal{S}, \mathcal{B}, \mathcal{O} \rightarrow \mathcal{E}$)}
The final stage generates detailed clinical examination results $\mathcal{E}$ based on the specific symptoms $\mathcal{S}$ observed in the previous stage, the patient profile $\mathcal{B}$, and the clinical criteria defined in $\mathcal{O}$. We enforce quantitative fidelity by instructing the model to generate specific metrics
% (e.g., tumor dimensions in millimeters, CT values in Hounsfield Units, biomarker concentration levels)
rather than vague descriptions. The final synthetic record is defined as $\mathcal{P} = \{\mathcal{B}, \mathcal{S}, \mathcal{E}\}$, where $\mathcal{O}$ serves as the latent structural scaffold. The complete synthetic patient record is provided in Appendix~\ref{app:case_study}.

% \paragraph{Step 3: Detailed Information Generation.}
% In the final phase, clinical examination results are generated based on the patient's symptoms and disease outline. A carefully selected one-shot prompt guides the generation of complex medical data while maintaining diversity. Each output is cross-referenced with prior information to ensure coherence throughout the synthetic patient record. After the mentioned steps, we can obtain a full synthetic patient record, as shown in Appendix~\ref{app:case_study}. Note that since our focus here is on the language models, hence only textual content without images is generated, which will be addressed in our future work. 
% More details on the generation and prompts are listed in Appendix~\ref{sec:prompt}.

% \vspace{-0.2em}
\subsection{Dual-Track Cognitive Memory System}
\label{subsec:memory_architecture}
\begin{figure}[t]
    \centering
    \includegraphics[width=\linewidth]{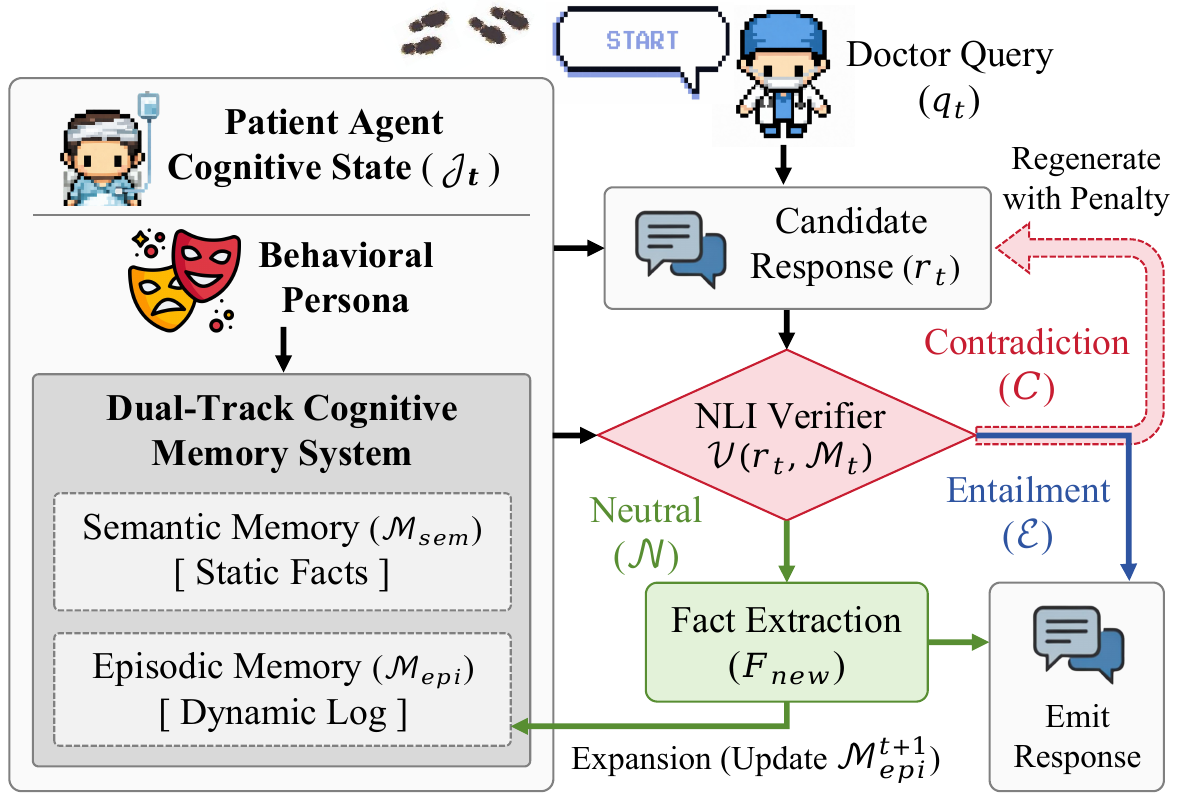}
    % \vspace{-15pt}
    \caption{\textbf{The Dual-Track Cognitive Memory System.} This module integrates static semantic memory ($\mathcal{M}_{sem}$) and dynamic episodic memory ($\mathcal{M}_{epi}$) to drive coherent dialogue. The \textit{NLI-Verifier} acts as a logical gatekeeper, evaluating candidate responses ($r_t$) against atomic memory ($\mathcal{M}_t$). By regenerating on Contradictions ($\mathcal{C}$), preserving state on Entailments ($\mathcal{E}$), and expanding on Neutral ($\mathcal{N}$) information, this closed-loop effectively resolves the \textit{\textbf{Stability-Plasticity Dilemma}}.}
    \label{fig:simulation_mechanism}
    % \vspace{-12pt}
\end{figure}

Bridging the gap between static patient records and dynamic clinical encounters requires transforming the record $\mathcal{P}$ into an embodied agent capable of coherent dialogue. A central challenge in this transformation is the \textit{\textbf{Stability-Plasticity Dilemma}}\footnote{\label{fn:dilemma}Originating in cognitive science and neural networks, the \textit{\textbf{Stability-Plasticity Dilemma}} refers to the computational trade-off between the ability of a system to retain existing knowledge (\textit{stability}) and its capacity to acquire new information without overriding valid prior memories (\textit{plasticity}). In our context, this manifests as the tension between rigid medical adherence and open-ended conversational adaptability.}~\cite{stability-plasticity-dilemma}: the agent must strictly adhere to the medical knowledge and memory facts (\textit{stability}) while naturally interacting for unscripted scenarios (\textit{plasticity}). 
% We resolve this via a \textbf{Dual-Track Cognitive Memory System} controlled by a Natural Language Inference \textit{(NLI)-Verifier}. 
To resolve this, we propose a Dual-Track Cognitive Memory System (Figure \ref{fig:simulation_mechanism}). 
Drawing upon the cognitive framework of \textit{declarative memory}~\cite{Riedel2015DeclarativeMemory}, we operationalize agent memory not as a monolithic buffer, but as a processing \textit{continuum} between two distinct levels of abstraction: \textit{semantic memory} for \textit{stability} and \textit{episodic memory} for \textit{plasticity}~\cite{Squire2004MemorySystems, Tulving2002EpisodicMemory, hu2025memoryageaiagents}.

% \subsubsection{Dual-Track Cognitive Memory System}
% \label{subsec:memory_architecture}

\paragraph{1) Semantic Memory ($\mathcal{M}_{sem}$):} 
In neuroscience, \textit{semantic memory} retains general factual knowledge independent of the specific occasion of acquisition~\cite{Squire2004MemorySystems}. In our framework, we instantiate $\mathcal{M}_{sem}$ using the generated patient record $\mathcal{P}$, serving as the ground-truth anchor to ensure consistency over time. To enable granular logical reasoning, we decompose the complex narrative record $\mathcal{P}$ into a discrete set of atomic propositions $\mathcal{M}_{sem} = \{f_1, f_2, \dots, f_n\}$, %Unlike monolithic context windows, this atomized storage 
preventing ``lost-in-the-middle'' phenomena during logical verification.

% \vspace{-0.2em}
\paragraph{2) Episodic Memory ($\mathcal{M}_{epi}$):} 
Complementing the semantic layer, \textit{episodic memory} stores personally experienced events associated with specific temporal contexts~\cite{Tulving2002EpisodicMemory}. We model $\mathcal{M}_{epi}$ as a dynamic log of the ongoing clinical dialogue. This track supports consistency by allowing the agent to mentally re-experience past events, and progressively aligning responses with the doctor's inquiry logic. Unlike the static $\mathcal{M}_{sem}$, this track captures the interaction histories as episodic traces~\cite{Zhong2024Memorybank, chhikara2025mem0buildingproductionreadyai}.
% \textit{what}, \textit{where}, and \textit{when} of the interaction trajectory.

% \subsubsection{Memory Evolution and Agent Persona}
% \paragraph{Consistency Validation via NLI} 
% \vspace{-0.2em}
\subsubsection{NLI-Driven Interactive Simulation}
To transform these memory structures into dynamic behavior, we define the agent cognitive state at turn $t$ as $\mathcal{J}_t = (\mathcal{M}_{sem} \cup \mathcal{M}_{epi}^t, \Psi)$. Here, $\Psi$ represents a behavioral persona (e.g., \textit{Reserved, Verbose, Upset}) injected following the taxonomy in \citet{patientpsi}, directly mapping to real communication barriers~\cite{lizée2024conversationalmedicalaiready}. As highlighted in recent surveys, a critical challenge is preventing hallucinations during the transformation of raw events into memory facts~\cite{tan2025reflective, tan-etal-2025-membench}. We address this via a \textit{NLI-Verifier} that enforces \textit{stability} by rejecting hallucinations, and enables \textit{plasticity} by integrating new non-conflicting details.
Importantly, the NLI-Verifier is a semantic and persona-consistency gatekeeper rather than a standalone numerical medical validator. Formalizable physiological, arithmetic, demographic, anatomical, and temporal constraints are handled by the deterministic validation layer described in Appendix~\ref{app:validity_rules}.

\paragraph{Persona and Logical Consistency}
At turn $t$, the patient agent generates a candidate response $r_t$ conditioned on the cognitive state $\mathcal{J}_t$ and doctor query $q_t$. We use an \textit{NLI-Verifier} $\mathcal{V}(r_t,\mathcal{M}_t,\Psi)$ to check both factual consistency and persona adherence, where $\mathcal{M}_t=\mathcal{M}_{sem}\cup\mathcal{M}_{epi}^t$. The verifier labels the relation between $r_t$ and each memory fact as entailment, contradiction, or neutral, and aggregates these labels using a strict priority rule: contradiction overrides entailment and neutrality. Any response contradicting semantic, episodic history, or persona traits is rejected and regenerated; grounded responses are emitted directly; and neutral but consistent information can update episodic memory. 

\paragraph{Memory Evolution Policy}
The agent memory evolves as a deterministic function of the verification outcome. We define the memory transition operator $\mathcal{M}_{t+1} \leftarrow \textsc{Update}(\mathcal{M}_t, r_t)$ conditioned on the verification result $\mathbb{V}(r_t, \mathcal{M}_t)$ as follows: %\vspace{-3pt}
% \textbf{1) Rejection (}$\mathbb{V} \to \mathcal{C}$; or \textit{persona violation}\textbf{):} The response violates factual constraints in $\mathcal{M}_t$ or deviates from persona $\Psi$ (\textit{violates stability constraint}). It is rejected and regenerated with explicit corrective instructions based on the detected contradiction or persona mismatch.
% \textbf{2) Invariance ($\mathbb{V} \to \mathcal{E}$):} The response $r_t$ is fully grounded in existing memory, and is emitted without modifying the memory state.
% \textbf{3) Expansion ($\mathbb{V} \to \mathcal{N}$):} The response updates new information that is logically consistent with $\mathcal{M}_t$ (\textit{representing plasticity}). Let $\mathcal{F}_{\text{new}}$ denote the set of newly extracted atomic facts from $r_t$.
% The \textit{episodic memory} is updated via monotonic expansion: $\mathcal{M}_{epi}^{t+1} \leftarrow \mathcal{M}_{epi}^t \cup \mathcal{F}_{\text{new}}$.
% \vspace{-2pt}
\begin{itemize}[left=0.3cm, itemsep=2pt, parsep=0pt]
    \item \textbf{Rejection (}$\mathbb{V} \to \mathcal{C}$; or \textit{persona violation}\textbf{):} The response violates factual constraints in $\mathcal{M}_t$ or deviates from persona $\Psi$ (\textit{violates stability constraint}). It is rejected and regenerated with explicit corrective instructions based on the detected contradiction or persona mismatch.
    \item \textbf{Invariance ($\mathbb{V} \to \mathcal{E}$):} The response $r_t$ is fully grounded in existing memory, and is emitted without modifying the memory state.
    \item \textbf{Expansion ($\mathbb{V} \to \mathcal{N}$):} The response updates new information that is logically consistent with $\mathcal{M}_t$ (\textit{representing plasticity}). Let $\mathcal{F}_{\text{new}}$ denote the set of newly extracted atomic facts from $r_t$.
    The \textit{episodic memory} is updated via monotonic expansion: $\mathcal{M}_{epi}^{t+1} \leftarrow \mathcal{M}_{epi}^t \cup \mathcal{F}_{\text{new}}$.
\end{itemize}

\section{Experimental Evaluation} 
\label{sec:experiments}

\begin{table*}[t]
\centering
\begin{adjustbox}{width=\textwidth}
\begin{tabular}{l|c|cccccc}
\toprule
\multirow{3}{*}{\textbf{Method}} & \multirow{3}{*}{\textbf{Backbone}} & \multicolumn{2}{c}{\textbf{Linguistic Quality}} & \multicolumn{2}{c}{\textbf{Semantic Diversity}} & \multicolumn{2}{c}{\textbf{Clinical Validity}} \\
\cmidrule(lr){3-4} \cmidrule(lr){5-6} \cmidrule(lr){7-8} 
& & \textbf{PPL} ($\downarrow$) & \textbf{Distinct-4} ($\uparrow$)& \makecell{\textbf{Self-}\\\textbf{BLEU}} ($\downarrow$) & \makecell{\textbf{Entity}\\\textbf{Diversity}} ($\uparrow$) & \makecell{\textbf{Consis}\\\textbf{-tency}} ($\uparrow$) & \makecell{\textbf{Complete}\\\textbf{-ness}} ($\uparrow$) \\
\midrule
\rowcolor{gray!10}\multicolumn{8}{c}{\textbf{Real-world Data Baselines}} \\
\addlinespace[2pt]
\color{gray} MIMIC-IV~\cite{mimic-iv} &\color{gray} -- & \color{gray} 1.50 & \color{gray} 10.99 & \color{gray} 85.89 & \color{gray} 4.15 & \color{gray} 92.73 & \color{gray} 45.44 \\
% \color{gray}ENCoDE~\cite{encode} \\
\color{gray}SCRIPT X2B8~\cite{SCRIPT-X2B8} & \color{gray} -- & \color{gray} 2.38 & \color{gray} 51.06 & \color{gray}74.75 & \color{gray}0.78 & \color{gray}96.12 & \color{gray}56.56 \\
\color{gray}CMA Base~\cite{CMA-Base} & \color{gray} -- & \color{gray}4.82 & \color{gray}83.63 & \color{gray}42.20 & \color{gray}27.02 & \color{gray}96.78 & \color{gray}91.35 \\
\midrule
\rowcolor{gray!10}\multicolumn{8}{c}{\textbf{Synthetic Data Baselines}} \\
\addlinespace[2pt]
Synthea~\cite{synthea} & Rule-based & 12.20 & 20.00 & 93.24 & 5.84 & 79.66 & 35.32 \\
LDP-GAN~\cite{LDP-GAN} & GAN & 10.86 & 30.70 & 91.18 & 1.10 & 85.89 & 46.40 \\
Avatar~\cite{Avatar} & FAMD + KNN & 9.03 & 48.43 & 70.42 & 3.74 & 95.22 & 42.84 \\
% Patient-$\Psi$~\cite{patientpsi} & & & & & & \\
MERA-Mistral~\cite{mera} & Mistral-7B-v0.3 & 3.88 & 50.43 & 77.61 & 6.18 & 75.84 & 69.20 \\
MERA-Llama~\cite{mera} & Llama-3-70B & \underline{3.51} & 50.57 & 77.09 & 8.27 & 82.15 & 97.00 \\
MERA-Qwen~\cite{mera} & Qwen-2.5-32B & \textbf{3.40} & 66.02 & 68.45 & 7.88 & 88.88 & 95.80 \\
\midrule
\rowcolor{gray!10}\multicolumn{8}{c}{\textbf{Ablation Baselines}} \\
\addlinespace[2pt]
\multirow{3}{*}{Direct (w/o Hierarchical Synthesis)} & GPT-5 & 9.19 & 61.83 & 59.16 & 21.69 & 70.76 & 75.75 \\
 & Gemini-2.5-Pro & 7.53 & 62.00 & 70.61 & 19.98 & 79.42 & 76.13 \\
 & Claude-Sonnet-4 & 9.16 & 65.89 & 61.83 & 21.78 & 71.35 & 61.50 \\
\midrule
\rowcolor{gray!20}
 & GPT-5 & 4.48$_{\text{ \textbf{\gain{-4.71}}}}$ & \textbf{76.51}$_{\text{ \textbf{\gain{+14.68}}}}$ & \textbf{52.45}$_{\text{ \textbf{\gain{-6.71}}}}$ & \textbf{23.72}$_{\text{ \textbf{\gain{+2.03}}}}$ & \textbf{99.12}$_{\text{ \textbf{\gain{+28.36}}}}$ & \textbf{99.88}$_{\text{ \textbf{\gain{+24.13}}}}$ \\
 \rowcolor{gray!20}
\textbf{Patient-Zero} &  Gemini-2.5-Pro & 4.23$_{\text{ \textbf{\gain{-3.30}}}}$ & \underline{74.15}$_{\text{ \textbf{\gain{+12.15}}}}$ & 57.64$_{\text{ \textbf{\gain{-12.97}}}}$ & \underline{23.36}$_{\text{ \textbf{\gain{+3.38}}}}$ & \underline{98.89}$_{\text{ \textbf{\gain{+19.47}}}}$ & \underline{98.91}$_{\text{ \textbf{\gain{+22.78}}}}$ \\
\rowcolor{gray!20}
& Claude-Sonnet-4 & 4.37$_{\text{ \textbf{\gain{-4.79}}}}$ & 71.35$_{\text{ \textbf{\gain{+5.46}}}}$ & \underline{55.67}$_{\text{ \textbf{\gain{-6.16}}}}$ & 22.55$_{\text{ \textbf{\gain{+0.77}}}}$ & 98.23$_{\text{ \textbf{\gain{+26.88}}}}$ & 98.47$_{\text{ \textbf{\gain{+36.97}}}}$ \\
\bottomrule
\end{tabular} 
\end{adjustbox}
\caption{Overall performance of Data Quality. Small \gain{gain} subscripts denote the performance gain compared to the ablation baseline using the same backbone, while  \textcolor{gray}{\textbf{real-world data}} serves as the reference baselines.}
\vspace{-8pt}
\label{tab:record_performance}
\end{table*}

% \vspace{-0.2em}
Our evaluation are structured around three core research questions (RQs):  
\textbf{(RQ1) Data Quality:} Does \textbf{Patient-Zero} synthesize records that are clinically consistent, diverse, and aligned with real-world distributions? 
\textbf{(RQ2) Interaction Fidelity:} Does our \textit{Dual-Track Cognitive Memory} resolve the \textit{Stability-Plasticity Dilemma}\textsuperscript{\ref{fn:dilemma}} in clinical dialogues?
\textbf{(RQ3) Downstream Utility:} Does Patient-Zero provide useful patient simulation for medical education, and does the generated data also support downstream medical reasoning model training?

% \vspace{-0.1em}
\subsection{Experimental Setup} 
\label{subsec:setup}
\paragraph{Datasets} 
We constructed a large-scale synthetic dataset comprising 60,000 patient records spanning six major clinical specialties: Cardiology, Gastroenterology, General Surgery, Neurology, Psychiatry, and Pulmonology. The dataset covers \textbf{98 distinct disease types}, ranging from common chronic conditions to critical acute emergencies. The distribution adheres to real-world epidemiological prevalence. Detailed disease and specialty statistics are provided in Appendix~\ref{app:data_construction}.

% \paragraph{Baselines} 
% To validate our framework, we employ distinct state-of-the-art baselines for each research question (see Appendix~\ref{app:baselines} for detailed configurations). 
% \textbf{1) Data Quality:} We evaluate our synthesized records against \textcolor{gray}{\textbf{real-world data}} (serving as Gold Standard), synthetic data, and ablation baselines.
% \textbf{2) Interaction Fidelity:} We compare our embodied agent with medical agent frameworks.

% \vspace{-0.1em}
\paragraph{Evaluation Metrics} 
A multi-facet protocol is used to assess the three research questions across distinct state-of-the-art baselines:
\textbf{1) Data Quality:} We evaluate the synthesized records across \textit{Linguistic Quality} (Perplexity, Distinct-4), \textit{Semantic Diversity} (Self-BLEU, Entity Diversity), and \textit{Clinical Validity} (Consistency, Completeness). Consistency measures guideline-referenced compatibility among clinical sections, and completeness measures required-section and field coverage.
\textbf{2) Interaction Fidelity:} We assess the agent interactive performance via \textit{Factual Fidelity} (Logical Consistency, Factual Recall), \textit{Behavioral Fidelity} (Persona Alignment, Stylistic Stability), \textit{Safety and Robustness} (Hallucination Rate, Inducibility Resistance).
\textbf{3) Downstream Utility:} Accuracy improvements on MedQA and MMLU, and also medical-student user studies. 
(see Appendix~\ref{app:baselines} for baseline configurations and Appendix~\ref{app:metrics} for metrics calculation methodologies).

% \vspace{-0.1em}
\subsection{(RQ1) Quality of Synthetic Data}
\label{subsec:record_quality}
% \paragraph{Overall Performance} 
\textbf{Patient-Zero} outperforms the real-world data approximations and synthetic baselines in Table~\ref{tab:record_performance}, consistently yielding substantial improvements in clinical validity ($>$98\%), demonstrating the generalizability across different backbone architectures.

% \vspace{-0.2em}
\paragraph{1) Breaking the Privacy-Utility Trade-off}
A critical challenge in patient data synthesis is the trade-off between privacy protection and data utility. Traditional synthetic baselines often suffer from mode collapse and clinical accuracy degradation due to privacy noise injection. In contrast, \textbf{Patient-Zero} effectively resolves this dilemma. By decoupling privacy from generation via \textit{ab initio} synthesis, we achieve near-perfect clinical validity (>98\%), while maintaining superior diversity.

\paragraph{2) Superiority over Direct Synthesis}
To validate the necessity of our \textit{Hierarchical Synthesis} framework, we compare \textbf{Patient-Zero} against its ablation baseline. Direct synthesis often yields outputs with high repetition and low clinical validity, as indicated by the \gain{gain} subscripts in Table~\ref{tab:record_performance}. This confirms that our hierarchical constraints effectively mitigate textual degeneration, enable models to generate non-stereotypical patient narratives.

\paragraph{3) Holistic Metric Balance} 
While MERA baselines achieve lower Perpelxity (PPL) at the cost of lexical richness, with repetitive generic medical phrasing. In contrast, \textbf{Patient-Zero} achieves significantly higher lexical diversity (+15.45 in Distinct-4 over MERA-Qwen), while does not degrade clinical validity, confirming that our data represent \textbf{\textit{rich, non-stereotypical clinical narratives rather than merely optimizing for syntactic predictability.}}

% \vspace{-0.2em}
\subsection{(RQ2) Fidelity of Interaction}
\begin{table*}[t]
\centering
% \vspace{-6pt}
\resizebox{\textwidth}{!}{
\begin{tabular}{l|cc|cc|cc}
\toprule
\multirow{3}{*}{\textbf{Method}} & \multicolumn{2}{c|}{\textbf{Factual Fidelity} (Cognitive)} & \multicolumn{2}{c|}{\textbf{Behavioral Fidelity} (Persona)} & \multicolumn{2}{c}{\textbf{Safety \& Robustness}} \\
\cmidrule(lr){2-3} \cmidrule(lr){4-5} \cmidrule(lr){6-7}
& \textbf{\makecell{Logical Con\\-sistency}($\uparrow$)} & \textbf{\makecell{Factual\\Recall}($\uparrow$)} & \textbf{\makecell{Persona\\Alignment}($\uparrow$)} & \textbf{\makecell{Stylistic\\Stability }($\uparrow$)} & \textbf{\makecell{Hallucination\\Rate}($\downarrow$)} & \textbf{\makecell{Inducibility\\Resistance}($\uparrow$)} \\
\midrule
\rowcolor{gray!10}\multicolumn{7}{c}{\textbf{State-of-the-art Medical Agents}} \\
\hspace{1em} EvoPatient~\cite{patient-agent-coevolution} & 87.49 & 9.78 & 80.50 & 32.95 & 5.01 & 94.87 \\
\hspace{1em} AI Hospital~\cite{fan2024aihospitalbenchmarkinglarge} & 97.95 & 16.68 & 78.80 & 29.33 & 6.67 & 86.36 \\
\hspace{1em} MediQ~\cite{mediq}  & 81.83 & 27.04 & 77.20 & 28.03 & 8.46 & 92.86 \\
\hspace{1em} Patient-$\Psi$~\cite{patientpsi} & 78.75 & 10.10 & 71.20 & 40.74 & 2.00 & 88.89 \\
\midrule
% \rowcolor{gray!10}\multicolumn{7}{l}{\textit{\textbf{Internal Baselines (Ablation)}}} \\
\rowcolor{gray!10}\multicolumn{7}{c}{\textbf{Architectural Ablations}} \\
\hspace{1em} Unstructured Narrative & 90.00 & 15.39 & 60.20 & 61.11 & 4.55 & 87.72 \\
\hspace{1em} Static Structured Schema & 93.18 & 14.41 & 67.20 & 64.60 & 2.73 & 79.90  \\
\hspace{1em} without \textit{Dual-Track Cog. Mem.} & 81.36 & 7.88 & 61.10 & 46.61 & 10.45 & 69.17 \\
\rowcolor{gray!20}
\textbf{Patient-Zero} & \textbf{100.00}$_{\text{ \textbf{\gain{+18.64}}}}$ & \textbf{33.27}$_{\text{ \textbf{\gain{+25.39}}}}$ & \textbf{83.70}$_{\text{ \textbf{\gain{+22.60}}}}$ & \textbf{75.50}$_{\text{ \textbf{\gain{+28.89}}}}$ & \textbf{1.82}$_{\text{ \textbf{\gain{-8.63}}}}$ & \textbf{97.22}$_{\text{ \textbf{\gain{+28.05}}}}$ \\

\bottomrule
\end{tabular}
}
\caption{Overall performance of Interaction Fidelity. Small \gain{gain} subscripts denote the absolute performance gain compared to the ablation baseline (without \textit{Dual-Track Cognitive Memory}).}
\label{tab:interaction_fidelity}
\vspace{-6pt}
\end{table*}

We operationalize Interaction Fidelity as the resolution of the \textit{Stability-Plasticity Dilemma}. We simulate multi-turn diagnostic sessions using the advanced reasoning capabilities of \textsc{Doctor-R1}~\cite{lai2025doctorr1masteringclinicalinquiry}, explicitly instructed to execute adversarial probing to stress-test the patient agents.

% We evaluate interaction fidelity in multi-turn diagnostic consultations rather than static record generation. Each Patient-Zero agent is initialized with the synthetic record as semantic memory and then queried by Doctor-R1 in an adversarial diagnostic protocol. The doctor agent performs symptom elicitation, leading-question probes, contradiction probes, and persona-sensitive follow-up questions. This setup directly tests whether the agent can preserve immutable clinical facts while incorporating non-conflicting conversational context.

% \vspace{-0.2em}
\paragraph{1) Mitigating the Stability-Plasticity Dilemma}
% As shown in Table~\ref{tab:interaction_fidelity}, 
Our framework effectively resolves the trade-off between clinical consistency and conversational fidelity (Table~\ref{tab:interaction_fidelity}),
% Our memory design outperforms the architecture baselines which suffer from "lost-in-the-middle" phenomenon.
surpassing state-of-the-art agent baselines across all metrics. The substantial gain of +18.64\% and +25.39\% in consistency and recall in the ablation confirms that our \textit{Dual-Track Cognitive Memory} is essential for long-horizon coherence.

% \vspace{-0.2em}
\paragraph{2) Mitigating Compliance Bias}
A critical vulnerability in medical simulation is \textit{compliance bias}, where patient agents hallucinate symptoms solely to align with the doctor suggestions. Our framework outperforms EvoPatient~\cite{patient-agent-coevolution} and Patient-$\Psi$~\cite{patientpsi} in both Hallucination Rate and Inducibility Resistance, while improving +28.05\% resistance and reduce 8.63\% hallucination over the ablation. This validates the efficacy of our \textit{Dual-Track Cognitive Memory} by proactively rejecting hallucinated responses.

\begin{table}[t]
\centering\vspace{0pt}
% \vspace{-6pt}
\small
\resizebox{0.85\linewidth}{!}{
\begin{tabular}{@{}lcc@{}}
\toprule
Model & MedQA & MMLU \\
\midrule
% Gemini-2.5-Flash & 61.50 & 20.50 \\
% Claude Sonnet 4 & 89.50 & 89.50 \\
% GPT-4.1 & 89.00 & 92.00 \\
% \midrule
\href{https://huggingface.co/m42-health/Llama3-Med42-8B}{\textcolor{darkblue}{Med42-v2-8B}} & 62.50 & 59.50 \\
% DoctorAgent-RL  & 58.00 & 72.50 \\
\href{https://huggingface.co/TsinghuaC3I/Llama-3-8B-UltraMedical}{\textcolor{darkblue}{UltraMedical-8B}} & 74.00 & 68.50 \\
\href{https://github.com/baichuan-inc/Baichuan-M2-32B}{\textcolor{darkblue}{Baichuan-M2-32B}} & 82.50 & 84.00\\
\midrule
Qwen3-8B (Base Model) & 61.00 & 69.50\\
Qwen3-8B + Real Data & 83.50 & \textbf{85.50} \\
\rowcolor{gray!20}
Qwen3-8B + \textbf{Patient-Zero} & \textbf{85.00} & 83.00 \\
\bottomrule
\end{tabular}}
\caption{Downstream Performance and Results.}
\label{tab:medqa_downstream}
\vspace{-10pt}
\end{table}

\subsection{(RQ3) Downstream Practical Utility} 
\label{subsec:downstream}

\begin{figure*}[t]
     \centering
    \includegraphics[width=\linewidth]{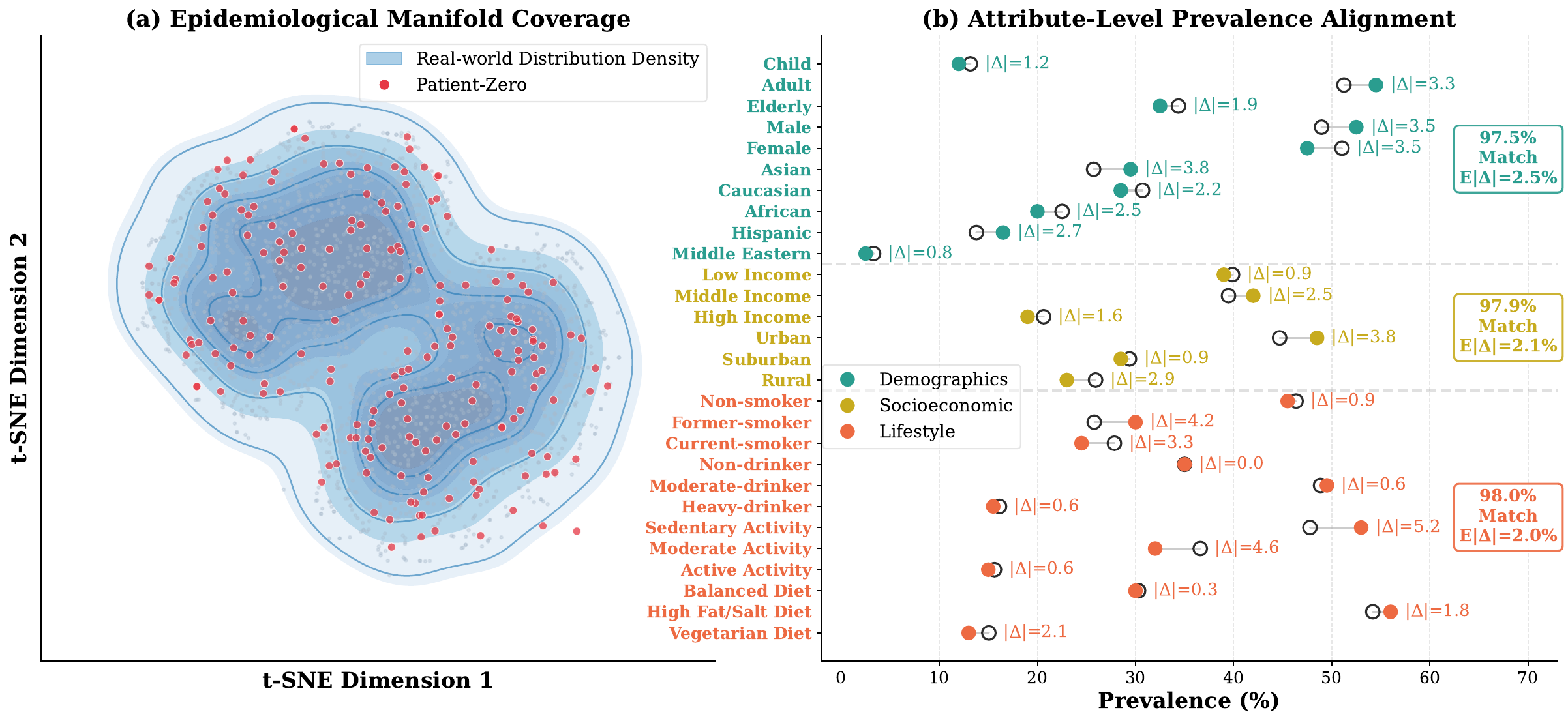}
    % \vspace{-21pt}
    \caption{\textbf{Holistic Epidemiological and Semantic Alignment.} \textbf{(a) Epidemiological Manifold Coverage:} The tight overlap of t-SNE visualization in the high-dimensional semantic space indicates that our synthetic data captures the comprehensive epidemiological manifold without mode collapse. \textbf{(b) Attribute-Level Prevalence Alignment:} Prevalence comparison across three dimensions. Hollow circles ($\circ$) represent real-world baseline; solid circles ($\bullet$) represent \textbf{Patient-Zero}; the connecting lines indicate alignment gaps. Minimal absolute differences demonstrate that Patient-Zero faithfully reconstructs complex real-world demographic and behavioral profiles.}
    \label{fig:data_distribution}
    % \vspace{-12pt}
\end{figure*}

To evaluate downstream utility, we trained Qwen3-8B~\cite{yang2025qwen3technicalreport} on \textbf{Patient-Zero} and assessed it on two clinical reasoning benchmarks: MedQA~\cite{jin2020diseasedoespatienthave} and MMLU~\cite{hendrycks2021measuringmassivemultitasklanguage}. As shown in Table~\ref{tab:medqa_downstream}, training on our synthetic data yields substantial gains over the base model (+24.0\% on MedQA; +14.5\% on MMLU). Under the same setup, a model trained on real data performs similarly, while model trained on \textbf{Patient-Zero} surpasses medical-LLM baselines, indicating that our synthetic data provides training utility comparable to real data (see Appendix~\ref{app:downstream} for detailed downstream training implementation).

To directly evaluate the educational utility of Patient-Zero, we conducted a user study with 16 medical students ranging from undergraduates to PhD candidates. Participants interacted with Patient-Zero agents in diagnostic consultations and rated the system on a 7-point Likert scale. In the medical student user study, Patient-Zero received high ratings across five educational dimensions: low-risk practice environment (6.31/7), teaching tool utility (6.06/7), response clarity (5.94/7), clinical judgment support (5.88/7), and communication skills improvement (5.69/7). 
The five-item questionnaire demonstrates good internal consistency with Cronbach's $\alpha=0.847$.
Relative to the neutral midpoint of the seven-point scale ($\mu_0=4$), all five dimensions exhibit large positive Cohen's $d$ and Hedges' $g$ effect sizes.
We therefore interpret these results as evidence of strongly positive perceived usability and educational value rather than as evidence of clinical effectiveness.
Detailed statistics are provided in Appendix~\ref{app:student_study}.
% These results support its utility as a scalable standardized patient for medical education (more details are listed in Appendix~\ref{app:student_study}).

% \vspace{-6pt}
% \vspace{-0.5em}
\subsection{Human Expert Evaluation} 
\label{subsec:human_eval}

\begin{figure}[h!]
    \centering
    % \vspace{-6pt}
    \includegraphics[width=\linewidth]{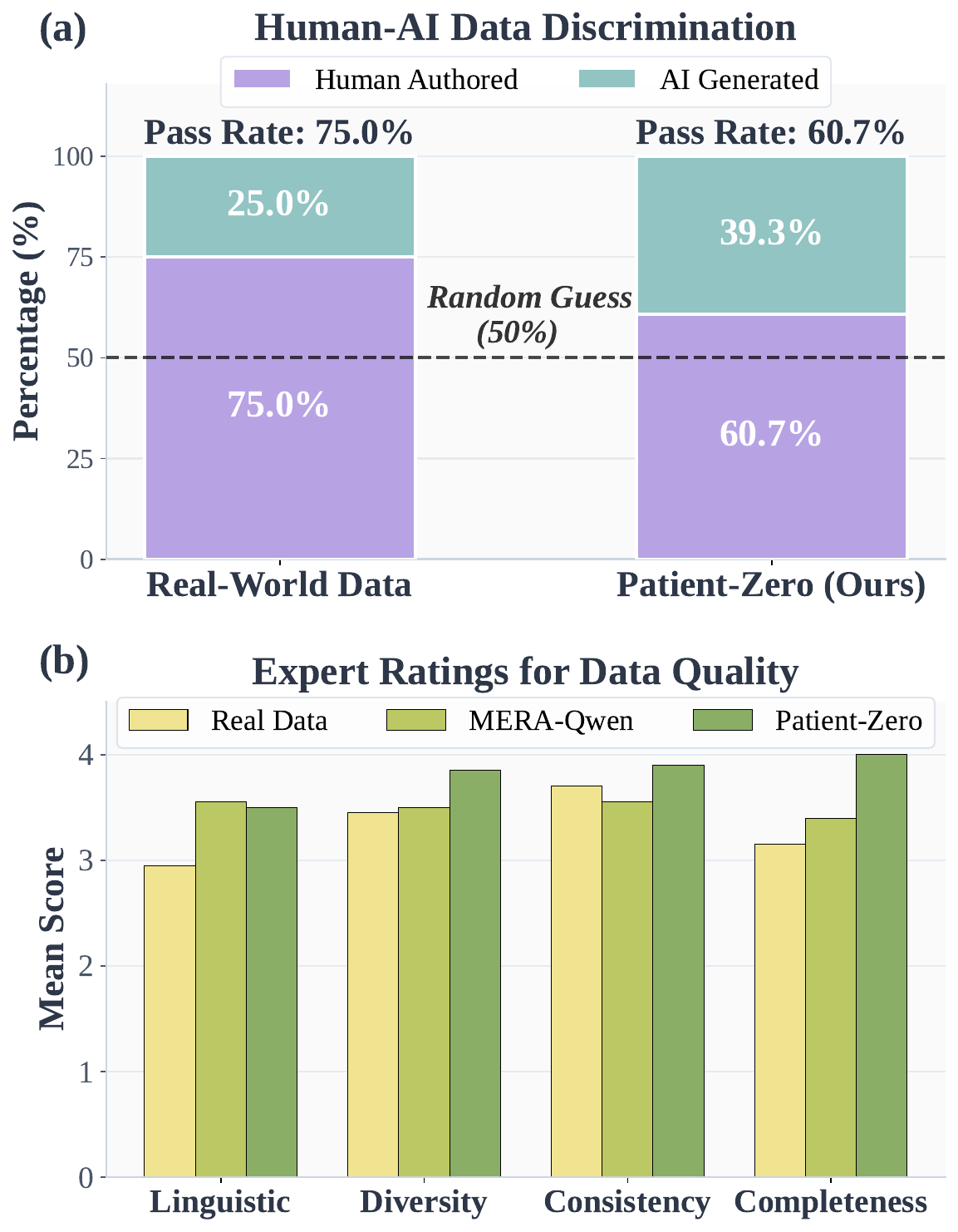}
    \vspace{-22pt}
    \caption{\textbf{Expert Evaluation Results.} (a) Senior licensed physicians judge discrimination between real and synthetic records. (b) Patient-Zero obtained the highest diversity, consistency, and completeness means.}
    \label{fig:human_eval_combined}
    \vspace{-12pt}
\end{figure}

To assess the clinical validity of our synthetic data, we employed senior licensed physicians from top-tier hospitals to evaluate \textbf{Patient-Zero} against real clinical data and baseline models. We conclude a key finding from their feedback: \textbf{\textit{real-world data often deviates from the ``ideal'' clinical standard}}, exhibiting issues such as missing fields, chaotic logic, or structural noise. 
% In contrast, our framework generates idealized clinical narratives with superior structural integrity. 
This section reports the results of the blind discrimination test and the multi-dimensional quality ratings. Detailed annotation protocols, qualitative case studies, and statistical breakdowns are provided in Appendix~\ref{app:human_evaluation_details}.

% \vspace{-6pt}
% \vspace{-0.1em}
\paragraph{Human-AI Data Discrimination}
Four physicians provided 112 authorship judgments and 120 quality ratings in a blinded discrimination test. As shown in Figure \ref{fig:human_eval_combined}(a), Patient-Zero and real records were judged human-authored in 60.7\% and 75.0\% of judgments respectively. 
This result reveals a critical insight: \textbf{\textit{physicians rely on a quality heuristic to distinguish human-authored content from AI generation.}} Experts tend to classify data with high completeness and logical flow as human-authored, while flagging records with noise, incompleteness, and chaotic formatting as AI-generated. Consequently, 25\% of real human-authored records were misclassified as AI due to their inherent flaws.
\textbf{Patient-Zero} shows alignment with the physicians' stereotype of a \textit{good human-authored patient record}, demonstrating only a slight perceptible difference between our synthetic data and real clinical records.
Annotator agreement was high (Krippendorff's $\alpha=0.93$).

% \vspace{-0.1em}
\paragraph{Expert Quality Ratings}
Licensed physicians evaluate data quality across four dimensions on a Likert scale. In Figure~\ref{fig:human_eval_combined}(b), Patient-Zero obtained the highest descriptive overall score ($3.81\pm0.76$), diversity ($3.85\pm0.75$), consistency ($3.90\pm0.85$), and completeness ($4.00\pm0.65$). Completeness was higher than MERA-Qwen~\cite{mera} synthetic baseline ($p=0.0257$) and real data~\cite{CMA-Base} ($p=0.0184$) under Mann-Whitney $U$ tests. These results strongly validate the findings in Table~\ref{tab:record_performance}, showing that our method delivers higher clinical quality visible to human experts.

\section{Analysis}
\label{sec:analysis}
% \vspace{-0.1em}  

\paragraph{Distribution Analysis}
We evaluate distributional alignment between Patient-Zero and selected epidemiological priors. Figure~\ref{fig:data_distribution}(a) shows broad manifold coverage, while Figure~\ref{fig:data_distribution}(b) shows close attribute-level alignment across demographic, socioeconomic, and lifestyle variables. Chi-square tests over marginal distributions show no significant deviations ($p>0.05$), and Fisher's combined test yields an aggregate $p$-value of 0.86 (Appendix~\ref{app:distribution_stats}). These results support alignment with selected marginal priors, but should not be interpreted as full equivalence to real-world clinical populations. We further test clinically meaningful joint dependencies in a 5,000-patient multivariate risk stratification analysis, including age--obesity, smoking--age, and alcohol--obesity co-risk patterns (see Appendix~\ref{app:multivariate-risk} for detailed elaboration).

\paragraph{Stage-wise Synthesis Ablation}
We conduct a stage-wise ablation using GPT-4.1 (version gpt-4.1-2025-04-14) across the six major clinical specialties: Cardiology, Gastroenterology, General Surgery, Neurology, Psychiatry, and Pulmonology. 
For each ablation setting, we generate 200 records and report mean performance across the six specialties.
Removing controlled sampling, deterministic rules, or semantic verification consistently degrades performance.
The largest consistency reduction occurs when semantic verification is removed.
Final-only verification also underperforms stage-wise verification, demonstrating the importance of controlling error propagation throughout the synthesis pipeline.

\begin{table}[t]
\centering
\small
\resizebox{\columnwidth}{!}{
\begin{tabular}{lrrrr}
\toprule
\textbf{Mode} & \textbf{S-BLEU}$\downarrow$ & \textbf{Ent. Div.}$\uparrow$ & \textbf{Cons.}$\uparrow$ & \textbf{Comp.}$\uparrow$ \\
\midrule
Hierarchical only & 26.64 & 54.06 & 80.46 & 97.72 \\
w/o controlled sampling & 17.86 & 58.44 & 86.26 & 90.49 \\
w/o deterministic rules & 17.92 & 57.56 & 88.60 & 95.87 \\
w/o semantic verifier & 21.40 & 57.72 & 84.92 & 96.36 \\
Final-only verification & 23.00 & 56.07 & 92.26 & 97.75 \\
\rowcolor{gray!20}
\textbf{Full Patient-Zero} & \textbf{15.27} & \textbf{63.33} & \textbf{95.86} & \textbf{99.55} \\
\bottomrule
\end{tabular}}
\caption{Stage-wise ablation of static synthesis.}
\vspace{-12pt}
\label{tab:stage_ablation}
\end{table}

\section{Conclusion}
\label{sec:conclusion}
% \vspace{-0.5em}
This paper presents \textbf{Patient-Zero}, an \textit{ab initio} framework connecting static patient-record synthesis with dynamic patient-agent simulation without using real patient records as generation seeds.
The \textit{Medically-Aligned Hierarchical Synthesis} module combines controlled attribute sampling, deterministic clinical constraints, and semantic verification to construct clinically coherent records aligned with selected epidemiological and medical-knowledge priors.
These records are subsequently instantiated as semantic memory within a \textit{Dual-Track Cognitive Memory System} for multi-turn patient simulation.
Our experiments demonstrate improvements in synthetic record quality, interaction fidelity, and downstream medical reasoning performance\.
Blinded evaluation by senior licensed physicians further shows strong descriptive ratings for consistency and completeness.
These results support Patient-Zero as a scalable framework for medical education and downstream model development, while we explicitly distinguish epidemiological-prior alignment and perceived human authorship from complete reproduction or certification of real-world clinical populations.

% Our study presents a multi-step generation framework that systematically enhances synthetic patient records through three core innovations: 1) a multi-step outline generation process guided by medical guidelines, 2) a real-time triplet evaluation mechanism with a memory update process for consistency checks, and 3) an adaptive conversational agent design supporting diverse interaction styles. Evaluations across medical specialties demonstrate measurable improvements in data accuracy, diversity, and MedQA accuracy compared to baseline methods. The system’s capacity to simulate context-aware dialogues provides new potentials for the usage of synthetic data in medical AI tasks. Our future research will focus on the integration of multimodal data to enable comprehensive medical record generation. Optimizing distribution-aware generation to help maintain statistical alignment between synthetic data and real-world distributions and demographic variations. 

\section*{Limitations}
\label{sec:limitations}
Methodology-wise, the primary limitation of our work is the \textit{idealization bias} inherent in our knowledge-anchored synthesis. While this structural integrity is favored by physicians, empirical studies on how this perfect data affects model robustness against real-world clinical noise shall be valuable. 
Scope-wise, we have not systematically extended the framework to \textit{multimodal} clinical data or validated the system in prospective clinical trials due to ethical and safety constraints. 
In future works, we plan to focus on constructing \textit{holistic virtual patients} that integrate visual and signal-based modalities. Additionally, we aim to theoretically uncover the relationship between synthetic data scale and downstream reasoning performance, specifically determining the optimal ratio of synthetic-to-real data required to maximize clinical utility without incurring hallucination risks.

\section*{Ethics Statement}

% This research presents a framework for generating synthetic patient records and is designed to mitigate privacy risks associated with real-world medical data. Since 
\textbf{Patient-Zero} generates data \textit{ab initio} based on abstract medical knowledge, hence no real patient data or private health information (PHI) is utilized or exposed in the development of this framework. This work strictly adheres to data privacy standards and poses no risk of re-identification. However, our approach is a research prototype intended for model training; the generated records are synthetic and should not be used as a substitute for real clinical data in healthcare decision-making without further validation.

\paragraph{Human Evaluation Protocol}
We conducted a human evaluation involving two senior licensed physicians from top-tier hospitals. 
These experts were tasked with a blinded discrimination test to distinguish between synthetic and real-world records, and also a multi-dimensional quality assessment. 
The evaluation was strictly double-blind. The annotators are unaware of the data sources and were instructed to evaluate based solely on defined qualitiative dimensions. We explicitly state that all participating expert annotators were fully informed about the study's purpose and data usage. They provided explicit consent via the user agreement on the annotation interface before commencing the evaluation tasks.

\paragraph{Annotator Welfare and Consent}
The participating physicians were fully informed about the purpose of the study and the tasks involved. All participation was voluntary, and the experts were compensated at a professional rate commensurate with their experience and local labor standards. No personal information regarding the evaluators was retained beyond their professional credentials required for the study.

\paragraph{Potential Risks}
We acknowledge potential risks inherent in synthetic medical data, including the possibility of hallucinated medical details or the amplification of biases present in the seed knowledge bases.
While our distribution analysis indicates high alignment with real-world statistics, synthetic cohorts may not fully capture rare anomalies or complex comorbidities found in organic populations. We release this dataset to facilitate privacy-safe research while emphasizing that models trained on this data should undergo safety testing before any deployment.

\paragraph{Data Governance and Compliance}
This study adheres to strict data governance protocols. We utilized the \textsc{MIMIC-IV} dataset~\cite{mimic-iv} solely for benchmarking and distribution comparison purposes. Access to MIMIC-IV was obtained under the \textit{PhysioNet Credentialed Health Data Use Agreement} (DUA). All authors accessing this data have completed the required CITI Program training in ``Data or Specimens Only Research''.
For open-source datasets and models, we strictly complied with their respective licenses and utilized them exclusively for academic research. 
Crucially, our proposed framework, \textbf{Patient-Zero}, generates synthetic records \textit{ab initio} from medical knowledge. Therefore, the released synthetic corpus contains no Personally Identifiable Information (PII) or Protected Health Information (PHI) derived from real patients, eliminating the risk of privacy leakage or re-identification.

\section*{Acknowledgments}
This work is supported by the Fundamental and Interdisciplinary Disciplines Breakthrough Plan of the Ministry of Education of China (No.JYB2025XDXM101),  Shenzhen Science and Technology Program (No. AI2026024), the Tam Shiu Charitable Trust, and Wuxi Research Institute of Applied Technologies Tsinghua University. Weizhi Ma is also supported by the Beijing Nova Program.

\bibliography{custom}
% \bibliographystyle{acl_natbib}

% \clearpage
\appendix
% \onecolumn

\section{LLM Usage Statement}
Throughout the completion of this work, the Large Language Model (LLM) was used solely for the purpose of refining sentences, improving grammatical accuracy and fluency during the manuscript writing process.

\section{Data Construction Details}
\label{app:data_construction}

\subsection{Hierarchical Attribute Taxonomy}
\label{app:attributes}
To ensure the synthesized population follows selected epidemiological priors encoded by the knowledge base, we employ a stratified sampling strategy based on the taxonomy detailed in Table \ref{tab:attribute_taxonomy}. This taxonomy comprehensively considers dimensions ranging from biological characteristics to the Social Determinants of Health (SDOH) \cite{who_sdoh}.

\begin{table*}[h!]
\centering
\small
\renewcommand{\arraystretch}{1.2}
\begin{tabularx}{\textwidth}{lX}
\toprule
\textbf{Dimension} & \textbf{Attributes and Permutations} \\
\midrule
\multirow{4}{*}{\makecell[l]{\textbf{Biological} \& \\\textbf{Demographic}}} 
& \textbf{Age Strata}: Child, Adult, Elderly \\
& \textbf{Biological Sex}: Male, Female \\
& \textbf{Physiological Status}: Pregnant, Non-pregnant, Post-menopausal \\
& \textbf{Ethnicity}: Asian, Caucasian, African American, Hispanic, Mixed, Other \\
\midrule
\multirow{2}{*}{\makecell[l]{\textbf{Socio-}\\\textbf{economic}}} 
& \textbf{Geography}: Urban (Metropolitan), Rural, Suburban; Specific Region Constraints \\
& \textbf{Socioeconomic Status}: Education Level, Occupation Type, Income Tier (Low/Middle/High) \\
\midrule
\multirow{4}{*}{\makecell[l]{\textbf{Behavioral}\\\& \textbf{Lifestyle}}} 
& \textbf{Substance Use}: Smoking (Never/Former/Current), Alcohol (None/Moderate/Heavy) \\
& \textbf{Diet \& Activity}: Dietary Pattern (Balanced/High-fat/High-salt); Activity Level (Sedentary/Moderate/Active) \\
& \textbf{Communication Style}: Plain, Upset, Verbose, Reserved, Tangent, Pleasing \\
& \textbf{Preferences}: Preference for Modern vs. Traditional Medicine \\
\bottomrule
\end{tabularx}
\caption{\textbf{Hierarchical Taxonomy of Patient Attributes.} This structure is used in the \textit{Patient-Zero} generation framework for attribute permutation to construct diverse patient records.}
\label{tab:attribute_taxonomy}
\end{table*}

\subsection{Disease Taxonomy Coverage}
\label{app:diseases}
The \textbf{Patient-Zero} dataset covers a wide spectrum of pathologies across six specialties, as shown in Table~\ref{tab:disease_list}. The selection ensures coverage of diverse clinical presentations, including infectious diseases, malignancies, chronic metabolic disorders, and acute trauma.

\begin{table*}[h!]
\vspace{12pt}
\centering
\small
\begin{tabularx}{\textwidth}{lX}
\toprule
\textbf{Specialty} & \textbf{Covered Diseases} \\
\midrule
\textbf{Cardiology} & Congenital Heart Disease, Hypertension, Pulmonary Adenocarcinoma, Esophageal Cancer, Fractured Rib, Atherosclerosis, Emphysema, Rheumatic Heart Disease, Acute Myocardial Infarction, Myocardial Ischemia, Heart Failure, Myocarditis, Arrhythmia, Hemothorax, Aortic Stenosis, Mediastinal Emphysema, Hypertensive Crisis, Coronary Artery Disease, Hyperlipidemia \\
\midrule
\textbf{General Surgery} & Hyperthyroidism, Mammary Hyperplasia, Cirrhosis, Corn, Hepatocellular Carcinoma, Thyroid Cancer, Colorectal Cancer, Breast Cancer, Gallbladder Polyps, Gallstones, Inguinal Hernia, Acute Appendicitis, Chronic Appendicitis, Pancreatitis, Anal Polyp \\
\midrule
\textbf{Neurology} & Stroke, Acute Myelitis, Tic Disorder, Cerebral Arteriosclerosis, Viral Meningitis, Hydrocephalus, Cerebral Ischemia, Leukoencephalopathy, Parkinson's Disease, Trigeminal Neuralgia, Vertigo, Ataxia, Vascular Dementia, Bacterial Meningitis, Alzheimer's Disease, Cerebral Hemorrhage, Cerebral Injury, Neurofibromatosis, Epilepsy, Cerebral Infarction, Migraine \\
\midrule
\textbf{Gastroenterology} & H. Pylori Infection, Chronic Gastritis, Gastric Ulcer, Hemorrhoids, GERD, IBS, Acute Gastroenteritis, Gastric Ptosis, Indigestion, Fatty Liver, Anal Fistula, Perianal Abscess, Anal Fissure, Rectal Prolapse, Rectal Polyp, Constipation, Liver Dysfunction \\
\midrule
\textbf{Pulmonology} & Asthma, Pulmonary Tuberculosis, Lung Cancer, Emphysema, Bronchitis, COVID-19 (Novel Coronavirus), Pneumonia, Influenza A, Influenza B, Mycoplasma Infection \\
\midrule
\textbf{Psychiatry} & Depression, ADHD, Auditory Hallucination, Anxiety Disorder, Schizophrenia, OCD, Insomnia, Bruxism, PTSD, Dissociative Disorders, Paranoid Personality Disorder, Social Anxiety Disorder, Anorexia, Eating Disorders, Phobia, Bipolar Disorder, Delusional Disorder \\
\bottomrule
\end{tabularx}
\vspace{-0.5em}
\caption{\textbf{Disease Taxonomy.} The dataset spans 98 distinct disease types across six major clinical specialties.}
\label{tab:disease_list}
\end{table*}

\subsection{Concrete Rules for the Validity Indicator}
\label{app:validity_rules}

In Table~\ref{tab:validity_constraint_groups}, the Validity Indicator $\mathbb{I}(\textsc{Valid}(\mathbf{v}))$ is implemented using explicit constraints grounded in the medical knowledge base.
A candidate patient state is retained only when all applicable constraint groups are satisfied.
Importantly, these constraints distinguish clinically intended abnormalities from invalid values.
A laboratory value outside the healthy reference range is therefore not rejected when the target disease is expected to produce the abnormality. Instead, the value is required to fall within its corresponding disease- and severity-conditioned interval. Table~\ref{tab:validity_rule_examples} shows the representative deterministic validation rules of different variables, with their corresponding acceptance conditions and failure actions.

\begin{table*}[t]
\centering
\small
\begin{tabular}{p{0.13\textwidth}p{0.25\textwidth}p{0.53\textwidth}}
\toprule
\textbf{Constraint Group} & \textbf{Purpose} & \textbf{Representative Checks} \\
\midrule
Attribute-range constraints &
Ensure sampled attributes lie within supported domains &
Valid age ranges; non-negative disease duration; physiologically possible height, weight, temperature, heart rate, and blood pressure. \\
\midrule
Demographic constraints &
Enforce compatibility between demographics and disease &
Pediatric conditions restricted to appropriate age ranges; age-associated diseases sampled according to disease-specific age priors. \\
\midrule
Anatomical constraints &
Prevent anatomically incompatible diagnoses &
Ovary-dependent conditions require the presence of ovaries; laterality-dependent findings must refer to an anatomically present structure. \\
\midrule
Disease-specific constraints &
Preserve core clinical characteristics &
Required symptoms or findings; prohibited symptom combinations; compatibility between disease subtype, severity, and examination findings. \\
\midrule
Comorbidity constraints &
Control incompatible disease combinations &
Mutually exclusive diagnoses and organ-function-dependent comorbidity restrictions. \\
\midrule
Temporal constraints &
Maintain valid longitudinal ordering &
Symptom onset precedes diagnosis; diagnosis precedes disease-specific treatment; treatment response cannot precede treatment initiation. \\
\midrule
Physiological constraints &
Ensure quantitative findings remain medically plausible &
Disease- and severity-conditioned ranges for blood counts, electrolytes, glucose, coagulation, inflammatory markers, and other measurements. \\
\midrule
Cross-variable constraints &
Preserve relationships among multiple fields &
BMI agrees with height and weight; anemia descriptions agree with hemoglobin measurements; symptom severity agrees with examination findings. \\
\bottomrule
\end{tabular}
\vspace{-0.5em}
\caption{Constraint groups used by the Validity Indicator.}
\label{tab:validity_constraint_groups}
\end{table*}

\begin{table*}[t]
\centering
\small
\begin{tabular}{p{0.18\textwidth}p{0.20\textwidth}p{0.33\textwidth}p{0.17\textwidth}}
\toprule
\textbf{Variable / Relation} & \textbf{Sampling Rule} & \textbf{Acceptance Condition} & \textbf{Failure Action} \\
\midrule
Age &
Sampled from $p(\mathrm{age}\mid d)$ &
Falls within the disease-supported interval &
Resample profile \\
\midrule
Disease duration &
Conditioned on visit type &
Positive and compatible with acute/chronic status &
Resample duration \\
\midrule
Height, weight, BMI &
Height and weight sampled jointly &
Reported BMI agrees with calculated BMI within tolerance $\epsilon$ &
Recalculate or regenerate \\
\midrule
Symptom onset and treatment &
Generated from ordered trajectory states &
Onset $<$ diagnosis $\leq$ treatment $<$ response/follow-up &
Regenerate trajectory \\
\midrule
Laboratory measurement &
Disease- and severity-conditioned interval &
Within the intended normal or pathological interval &
Resample measurement \\
\midrule
Anatomy-dependent diagnosis &
Conditioned on recorded anatomy &
Required anatomy is present &
Reject profile \\
\midrule
Follow-up encounter &
Conditioned on treatment history &
Prior diagnosis and treatment are present &
Regenerate encounter type \\
\bottomrule
\end{tabular}
\vspace{-0.5em}
\caption{Representative deterministic validation rules.}
\label{tab:validity_rule_examples}
\end{table*}

\subsection{NLI-Verifier and Failure Modes}
\label{app:nli_verifier}

The semantic verifier is implemented as a frozen inference-time model without task-specific fine-tuning. Detailed configuration the NLI-Verifier is listed in Table~\ref{tab:nli_configuration}.

\begin{table}[t]
\centering
\small
\begin{tabular}{lp{0.55\linewidth}}
\toprule
\textbf{Component} & \textbf{Specification} \\
\midrule
Model & GPT-4.1-2025-04-14 \\
\midrule
Training & Frozen; no task-specific fine-tuning \\
\midrule
References & Structured patient state, previous-stage outputs, and relevant medical guidelines \\
\midrule
Decision labels & Consistent / Neutral / Contradictory \\
\midrule
Checked relations & Cross-section consistency, temporal order, treatment--response compatibility, patient identity, and persona adherence \\
\midrule
Failure action & Regenerate the contradictory section, up to $N=3$ attempts \\
\midrule
Decoding & Temperature $=0.0$, top-$p=1.0$ \\
\bottomrule
\end{tabular}
\vspace{-0.5em}
\caption{Configuration of the NLI-Verifier.}
\label{tab:nli_configuration}
\end{table}

\paragraph{Decision Protocol}
The verifier does not use a learned calibration procedure.
Instead, we apply fixed label definitions: a section is classified as contradictory only when it conflicts with explicit patient facts, previously generated clinical content, or guideline-grounded constraints.
Cases lacking sufficient evidence are classified as neutral rather than contradictory.
This conservative decision rule reduces false contradiction judgments in clinically ambiguous cases.

\paragraph{Persona and Nuanced Clinical Entailment}
Persona adherence is evaluated against the semantic memory containing demographic identity, medical history, psychological traits, and communication style.
Nuanced clinical entailment is evaluated using the structured disease state and relevant medical-knowledge evidence as references.

\paragraph{Failure Modes}
The verifier may still fail on implicit persona relations, subtle temporal dependencies, and numerical trends.
For example, in an aortic stenosis case, a patient response reporting prior aortic stenosis while denying heart failure was incorrectly classified as contradictory against a record specifying no prior heart failure and no known congenital valvular disease.
This error reflects conflation of related but clinically distinct concepts, since aortic stenosis may be acquired or degenerative and does not imply congenital valvular disease.
We therefore do not treat the semantic verifier as exhaustive physiological certification.

\section{Experimental Configuration and Implementation}
\label{app:experimental_setup}

\subsection{Baselines Details}
\label{app:baselines}
We categorize our baselines into two distinct groups to strictly align with the evaluation tasks reported in Table~\ref{tab:record_performance} and Table~\ref{tab:interaction_fidelity}.

\paragraph{Baselines for Data Quality (RQ1)}
To evaluate the static quality of generated patient records ($\mathcal{P}$), we compare against four categories of baselines:

\begin{itemize}[left=0pt, itemsep=2pt, parsep=0pt]
    \item \textbf{Real-world Data Baselines:} We employ de-identified clinical records from three sources to serve as the reference baselines for distribution alignment: \textbf{1) MIMIC-IV}~\cite{mimic-iv}: Large-scale ICU electronic health records. \textbf{2) CMA Base}~\cite{CMA-Base}: A comprehensive clinical medical association dataset representing general practice distributions.

    \item \textbf{Traditional Synthetic Baselines:} Established non-LLM approaches representing previous generation paradigms: \textbf{1) Synthea} \cite{synthea}: A rule-based engine simulating patient lifespans and medical history. \textbf{2) LDP-GAN} \cite{LDP-GAN}: A Generative Adversarial Network incorporating Local Differential Privacy. \textbf{3) Avatar} \cite{Avatar}: A hybrid method combining Factor Analysis of Mixed Data (FAMD) with k-Nearest Neighbors (KNN).

    \item \textbf{LLM-based Synthetic Baselines}: We evaluate the \textbf{MERA}~\cite{mera} framework of different open-source backbones to benchmark generation capabilities, including MERA-Mistral (Mistral-7B-v0.3), MERA-Llama (Llama-3-70B), and MERA-Qwen (Qwen-2.5-32B).

    \item \textbf{Ablation Baselines:} To validate the necessity of our \textit{Hierarchical Synthesis}, we compare our framework against an approach without hierarchical constraints, using proprietary models: \textbf{GPT-5} (gpt-5-2025-08-07), \textbf{Gemini-2.5-Pro}, and \textbf{Claude-Sonnet-4} (claude-sonnet-4-20250514).
\end{itemize}

\paragraph{Baselines for Interaction Fidelity (RQ2)}
To evaluate the dynamic performance of the patient agent during medical consultations, we compare against state-of-the-art agents and internal architectural ablations:

\begin{itemize}[left=0pt, itemsep=2pt, parsep=0pt]
    \item \textbf{State-of-the-art Medical Agents:} \textbf{1) EvoPatient} \cite{patient-agent-coevolution}: Uses evolutionary algorithms to iteratively optimize patient profiles for diagnostic realism. \textbf{2) AI Hospital} \cite{fan2024aihospitalbenchmarkinglarge}: A multi-agent framework utilizing role-specific prompts for simulation. \textbf{3) MediQ} \cite{mediq}: A framework emphasizing accurate clinical reasoning and query response. \textbf{4) Patient-$\Psi$} \cite{patientpsi}: A cognitive modeling approach focusing on psychological and behavioral fidelity.

    \item \textbf{Architectural Ablations:} To isolate the contributions of our specific components: \textbf{1) Unstructured Narrative:} The agent utilizes raw text memory without structured decomposition. \textbf{2) Static Structured Schema:} The agent uses structured records but lacks dynamic episodic updates. \textbf{w/o Dual-Track Cognitive Memory:} The full framework but with the \textit{NLI-Verifier} loop removed, accepting all generated responses without consistency checks.
\end{itemize}

\paragraph{Baselines for Downstream Clinical Utility (RQ3)}
To assess the utility of our synthetic data in training downstream clinical reasoning models, we compare against leading open-source medical LLMs and real-data training setups:

\begin{itemize}[left=0pt, itemsep=2pt, parsep=0pt]
    \item \textbf{State-of-the-art Medical LLMs:} \textbf{1) Med42-v2-8B}~\cite{med42v2}: A specialized clinical LLM fine-tuned on extensive medical instruction datasets, serving as a strong domain-specific baseline. \textbf{2) UltraMedical-8B}~\cite{zhang2024ultramedical}: A state-of-the-art medical model enhanced with large-scale biomedical literature and medical knowledge. \textbf{3) Baichuan-M2-32B} \cite{m2team2025baichuanm2scalingmedicalcapability}: A large-scale bilingual medical foundation model demonstrating superior performance on clinical reasoning tasks.
    
    \item \textbf{Control Group Settings:} \textbf{1) Qwen3-8B (Base Model)}~\cite{yang2025qwen3technicalreport}: The unmodified foundation model, serving as the zero-shot baseline to measure the raw capability gain. \textbf{2) Qwen3-8B + Real Data:} The base model fine-tuned on real-world clinical records under identical experimental settings. This serves as the \textit{ceiling performance} reference to verify if synthetic data can match the supervision quality of organic data.
\end{itemize}

\subsection{Evaluation Metrics Definition}
\label{app:metrics}

We provide detailed definitions for the metrics used to evaluate data quality and interaction fidelity.

\paragraph{(RQ1) Data Quality Assessment}
To validate the synthesized patient records $\mathcal{P}$, we utilize three categories of metrics:
\begin{itemize}[left=0pt, itemsep=2pt, parsep=0pt]
    \item \textbf{Linguistic Quality:} We measure the fluency and lexical richness of the generated text. \textit{Perplexity (PPL)} quantifies the model's uncertainty, with lower scores indicating greater fluency. Instead of using generic open-domain models, we employ a state-of-the-art specialized medical LLM, \textbf{Baichuan-M2-32B}~\cite{m2team2025baichuanm2scalingmedicalcapability} to ensure scores reflect clinical semantic coherence rather than generic syntactic probability. \textit{Distinct-4} calculates the ratio of unique 4-grams, serving as a proxy for lexical diversity and the absence of repetitive degeneration. This ensures that the perplexity score reflects \textbf{clinical semantic coherence} and domain-specific plausibility rather than mere syntactic probability. A lower PPL under this model indicates that the generated record aligns well with professional medical corpus distributions.
    \item \textbf{Semantic Diversity:} To ensure \textbf{Patient-Zero} covers the ``long tail'' of patient distributions without mode collapse, we report \textit{Self-BLEU}, where lower scores indicate higher diversity between generated samples. Additionally, \textit{Entity Diversity} measures the unique count of medical entities recognized by a biomedical NER tagger, reflecting the clinical richness of the dataset.
    \item \textbf{Clinical Validity:} We employ GPT-4.1 (version gpt-4.1-2025-04-14) as a verified evaluator to assess medical alignment. \textit{Consistency} measures the logical coherence between symptoms, medical history, and diagnosis. \textit{Completeness} evaluates whether the record contains all necessary clinical sections (e.g., HPI, Physical Exam) in medical knowledge sources.
\end{itemize}

\paragraph{(RQ2) Interaction Fidelity Assessment}
To evaluate the agent's performance during dynamic consultations, we define metrics across three dimensions aligned with our Dual-Track architecture:
\begin{itemize}[left=0pt, itemsep=2pt, parsep=0pt]
    \item \textbf{Factual Fidelity (Cognitive):} Assesses the stability of the semantic memory $\mathcal{M}_{sem}$. \textit{Logical Consistency} is the percentage of agent responses that do not logically contradict the ground-truth patient record, verified via an NLI model. \textit{Factual Recall} reports the F1 score of key medical facts retrieved during the conversation compared to the static patient record.
    
    \item \textbf{Behavioral Fidelity (Persona):} Evaluates adherence to the injected persona $\Psi$. \textit{Persona Alignment} is reported as a normalized percentage score, measuring the extent to which the agent's responses exhibit the specific behavioral traits as assessed by expert evaluation. \textit{Stylistic Stability} measures the consistency of linguistic style features across multi-turn interactions.
    
    \item \textbf{Safety \& Robustness:} \textit{Hallucination Rate} quantifies the frequency of fabricating non-existent symptoms or medical history. \textit{Inducibility Resistance} measures the agent's safety capability to reject ``leading questions'' from the doctor (e.g., coercing the patient to admit a symptom they do not possess), which is critical for preventing misdiagnosis in medical simulation.
\end{itemize}

\subsection{Downstream Model Training}
\label{app:downstream}

\paragraph{Framework and Hardware}
We conducted all downstream training experiments on a computational node equipped with 8 NVIDIA A100 80GB GPUs. Our implementation is built upon the \textsc{VeRL} framework~\citep{Sheng_2025}, utilizing efficient distributed training strategies. To optimize inference throughput during the rollout phase, we integrated the vLLM engine. The policy optimization was performed using the Group Relative Policy Optimization (GRPO)~\cite{shao2024deepseekmathpushinglimitsmathematical} algorithm, a state-of-the-art method for stabilizing reinforcement learning in language models.

\paragraph{Model Configurations}
We utilized the Qwen3-8B~\cite{yang2025qwen3technicalreport} instruction-tuned model as the backbone for both the Actor (Policy) and Critic networks. To manage memory efficiency while maintaining precision, the models were loaded in \texttt{bfloat16} format. Gradient checkpointing was enabled to reduce memory footprint. For the optimization process, we employed separate learning rates for the actor and critic to ensure stable convergence: the Actor was trained with a learning rate of $1 \times 10^{-6}$, while the Critic utilized a higher learning rate of $1 \times 10^{-5}$. We set the KL divergence coefficient ($\beta_{KL}$) to $0.001$ to prevent the policy from deviating excessively from the reference model.

\paragraph{Training Data and Hyperparameters}
The training was conducted on the \textbf{Patient-Zero} synthetic dataset. We configured a global training batch size of 128, distributed across GPUs with a micro-batch size of 4 per device. The generation configuration allowed for a maximum prompt length of 2048 tokens and a maximum response length of 2048 tokens, ensuring sufficient context window for complex clinical reasoning. During the rollout phase, we sampled $n=5$ trajectories per prompt to robustly estimate the advantage function. The entire training process was monitored using Weight \& Biases (Wandb)\footnote{\url{https://wandb.ai/site}}, with a reward clipping mechanism applied to maintain training stability.

\paragraph{Benchmark}
To rigorously assess whether the specialized training on synthetic patient data induces \textit{catastrophic forgetting} of general medical knowledge, we extended our evaluation to two gold-standard multiple-choice benchmarks. The evaluation protocol was standardized on a randomly sampled subset of 200 questions from the \textbf{US English} test sets for MedQA dataset~\citep{jin2020diseasedoespatienthave} and MMLU~\citep{hendrycks2021measuringmassivemultitasklanguage} of medical topics.

\subsection{Medical Student User Study Details}
\label{app:student_study}

\begin{table*}[t]
\centering
\small
\begin{tabular}{p{0.35\textwidth}ccc}
\toprule
\textbf{Dimension} &
\textbf{Mean $\pm$ SD} &
\textbf{Cohen's $d$} &
\textbf{Hedges' $g$} \\
\midrule
Low-risk practice environment &
$6.31\pm1.25$ & 1.85 & 1.76 \\

Teaching tool utility &
$6.06\pm1.18$ & 1.75 & 1.66 \\

Response clarity &
$5.94\pm1.18$ & 1.64 & 1.56 \\

Clinical judgment support &
$5.88\pm1.26$ & 1.49 & 1.41 \\

Communication skills improvement &
$5.69\pm1.35$ & 1.25 & 1.18 \\
\bottomrule
\end{tabular}
\vspace{-0.5em}
\caption{\textbf{Medical Student User Study.} Effect sizes are calculated relative to the neutral midpoint $\mu_0=4$ of the seven-point Likert scale.}
\label{tab:student_effect_size}
\end{table*}

The questionnaire demonstrates good internal consistency with Cronbach's $\alpha=0.847$.
Since participants are survey respondents rather than annotators assigning labels to the same clinical cases, questionnaire internal consistency is the relevant reliability measure.

We calculate one-sample Cohen's $d$ relative to the neutral midpoint:
\begin{equation}
d=\frac{\bar{x}-\mu_0}{s},
\qquad \mu_0=4.
\end{equation}
Given the modest sample size ($n=16$), we additionally report the small-sample-corrected Hedges' $g$:
\begin{equation}
g=\left(1-\frac{3}{4(n-1)-1}\right)d.
\end{equation}
All five dimensions exhibit large positive effect sizes relative to the neutral midpoint.

\begin{table}[t]
\centering
% \small
\resizebox{0.4\textwidth}{!}{
\begin{tabular}{lc}
\toprule
\textbf{Hyperparameter} & \textbf{Value} \\
\midrule
\rowcolor{gray!10}
\multicolumn{2}{c}{\textit{General Settings}} \\
Base Model & Qwen3-8B \\
Algorithm & GRPO \\
Precision & bfloat16 \\
Number of GPUs & 8 \\
\midrule
\rowcolor{gray!10}
\multicolumn{2}{c}{\textit{Batch \& Rollout}} \\
Global Batch Size & 128 \\
PPO Mini-Batch Size & 64 \\
Micro-Batch Size (Per GPU) & 4 \\
Rollout Samples ($N$) & 5 \\
\midrule
\rowcolor{gray!10}
\multicolumn{2}{c}{\textit{Optimization}} \\
Actor Learning Rate & $1 \times 10^{-6}$ \\
Critic Learning Rate & $1 \times 10^{-5}$ \\
KL Coefficient ($\beta_{KL}$) & 0.001 \\
Max Prompt Length & 2048 \\
Max Response Length & 2048 \\
\bottomrule
\end{tabular}}
\vspace{-0.5em}
\caption{Hyperparameters for Downstream Training.}
\label{tab:hyperparams_downstream}
\end{table}

\section{Statistical Distribution Analysis}
\label{app:distribution_stats}
To rigorously validate the claim of \textit{Real-World Distribution Alignment}, we define the statistical framework used for the \textbf{Chi-square Goodness-of-Fit test} ($\chi^2$). We compared the generated \textbf{Patient-Zero} cohort ($N=200$) against ground-truth epidemiological priors derived from our Knowledge Base (Monte Carlo reference, $N=2000$).

\begin{definition}[Null Hypothesis for Distribution Alignment]
Let $\hat{\mathcal{P}}$ be the empirical distribution of the synthetic patient attributes and $\mathcal{P}^*$ be the real-world prior distribution. The null hypothesis $H_0$ posits that there is no statistically significant difference between the two distributions:
\begin{equation*}
    H_0: \hat{\mathcal{P}} = \mathcal{P}^*
\end{equation*}
We adopt a significance level of $\alpha = 0.05$. A $p$-value $> 0.05$ indicates a failure to reject $H_0$, confirming alignment.
\end{definition}

\paragraph{Hypothesis Testing Results}
Table~\ref{tab:chisquare_results} presents the results of the Chi-square Goodness-of-Fit tests across nine epidemiological features. 
We observe that \textbf{Patient-Zero} successfully passed the alignment test for all nine features ($100\%$ success rate). 
Notably, complex lifestyle attributes, which are typically prone to generation bias in LLMs, showed exceptionally high conformity: \textit{Alcohol Consumption} ($p=0.97, \chi^2=0.07$) and \textit{Dietary Habits} ($p=0.71, \chi^2=0.69$). 
The \textit{Fisher's combined probability test} yielded a global $p$-value of \textbf{0.86}, providing strong evidence that the synthetic cohort is statistically align with the real-world population.

\begin{table*}[h]
\centering
% \small
\caption{\textbf{Chi-Square Goodness-of-Fit Test Results.} The high $p$-values across all dimensions (especially Lifestyle attributes like Alcohol Consumption) strongly support the acceptance of $H_0$.}
\label{tab:chisquare_results}
\resizebox{0.7\textwidth}{!}{
\begin{tabular}{llccc}
\toprule
\textbf{Dimension} & \textbf{Feature} & \textbf{$\chi^2$ Statistic} & \textbf{$p$-Value} & \textbf{Result} \\
\midrule
\multirow{3}{*}{\textbf{Demographics}} 
& Age Group & 0.82 & 0.66 & \textcolor{teal}{Aligned} \\
& Biological Sex & 1.00 & 0.32 & \textcolor{teal}{Aligned} \\
& Ethnicity & 3.40 & 0.49 & \textcolor{teal}{Aligned} \\
\midrule
\multirow{2}{*}{\textbf{Socioeconomic}} 
& Income Level & 0.62 & 0.73 & \textcolor{teal}{Aligned} \\
& Geographic Location & 1.37 & 0.51 & \textcolor{teal}{Aligned} \\
\midrule
\multirow{4}{*}{\textbf{Lifestyle}} 
& Smoking Status & 2.21 & 0.33 & \textcolor{teal}{Aligned} \\
& Alcohol Consumption & \textbf{0.07} & \textbf{0.97} & \textcolor{teal}{Aligned} \\
& Physical Activity & 2.35 & 0.31 & \textcolor{teal}{Aligned} \\
& Dietary Habits & 0.69 & 0.71 & \textcolor{teal}{Aligned} \\
\midrule
\multicolumn{2}{l}{\textit{\textbf{Fisher Combined Statistic}}} & \textit{-} & \textbf{\textit{0.86}} & \textbf{\textit{Global Match}} \\
\bottomrule
\end{tabular}}
\end{table*}

\paragraph{Effect Size Analysis}
To ensure that the lack of statistical significance is due to genuine distributional similarity rather than insufficient sample power, we computed effect size metrics including \textit{Cramér's V} and \textit{Total Variation Distance (TVD)}.
\begin{itemize}[left=0pt]
    \item \textbf{Cramér's V:} The average Cramér's V across all features is \textbf{0.054}, with a maximum of 0.077 (Physical Activity). According to Cohen's guidelines~\cite{cohen1988statistical}, a value $V < 0.10$ indicates a \textit{\textbf{negligible effect size}}, confirming that the deviation between synthetic and real distributions is minimal.
    \item \textbf{Total Variation Distance (TVD):} The average TVD is \textbf{0.035}, implying that the synthetic probability mass function deviates from the ground truth by an average of only 3.5\% across all categories.
\end{itemize}

\paragraph{Granular Category Alignment}
We further examined the alignment at the granular category level. For instance, in the \textit{Ethnicity} feature, the synthetic distribution for minority groups closely mirrors the expected priors (e.g., Hispanic: Observed 17.0\% vs. Expected 14.3\%, $\Delta=+2.7\%$). Similarly, in \textit{Socioeconomic Status}, the Low Income group representation (39.0\%) is nearly identical to the ground truth (39.9\%). This granular fidelity ensures that \textbf{Patient-Zero} preserves the diversity of underrepresented subpopulations, mitigating the risk of mode collapse often seen in synthetic medical data.

\subsection{Multivariate Risk Stratification}
\label{app:multivariate-risk}

Beyond marginal epidemiological alignment, we further evaluate whether Patient-Zero preserves clinically meaningful joint dependencies among demographic, behavioral, and disease-risk attributes. We conduct a multivariate risk stratification analysis on a synthetic cohort of 5,000 patients. Rather than testing isolated attributes independently, this analysis examines whether clinically established risk amplification patterns emerge when multiple risk factors co-occur.

We consider three clinically motivated dependency structures: age and obesity for hypertension risk, smoking and age for lung disease risk, and alcohol consumption and obesity for fatty liver or liver dysfunction risk. For each scenario, we define a low-risk reference group and a high-risk exposure group according to the generated patient attributes, and compute disease prevalence and relative risk (RR) between the two groups.

For hypertension, Patient-Zero reproduces the expected synergistic risk pattern between age and obesity. The prevalence of hypertension is lowest in the low-risk group, consisting of young or non-obese patients, at 3.71\%. In contrast, the old-obese group shows a substantially higher prevalence of 23.64\%, corresponding to a relative risk of 6.37. This indicates that the generated cohort captures the amplified cardiovascular risk associated with the co-occurrence of aging and obesity.

For lung disease, Patient-Zero captures the cumulative exposure effect of smoking history. Young smokers show a lung-disease prevalence of 12.24\%, whereas old smokers show a markedly higher prevalence of 46.11\%, corresponding to a relative risk of 3.77. This pattern is consistent with the clinical expectation that smoking-related respiratory risk increases with longer exposure duration and aging.

For fatty liver or liver dysfunction, Patient-Zero captures the combined metabolic and alcohol-related risk structure. The baseline group, consisting of lean non-drinkers, shows a prevalence of 6.78\%. In contrast, obese drinkers show a much higher prevalence of 61.69\%, corresponding to a relative risk of 9.10. This result indicates that the generated profiles preserve clinically plausible co-risk between obesity-related metabolic burden and alcohol consumption.

Overall, these results show that Patient-Zero captures clinically meaningful co-occurrences and risk amplification patterns beyond marginal demographic alignment. They extend the marginal alignment analysis by testing whether generated patient profiles preserve multivariate clinical dependencies rather than merely matching isolated attribute distributions.

\section{Human Expert Evaluation Details}
\label{app:human_evaluation_details}

\begin{table*}[t]
\vspace{12pt}
\centering
\small
\begin{tabular}{llcccccc}
\toprule
\textbf{Source} & \textbf{Metric} & \textbf{N} & \textbf{Mean} & \textbf{SD} & \textbf{Median} & \textbf{95\% CI} \\
\midrule
\multirow{4}{*}{Patient-Zero} 
& Linguistic Quality & 10 & 2.40 & 0.52 & 2.0 & [2.03, 2.77] \\
& Information Richness & 10 & 3.40 & 0.70 & 3.5 & [2.90, 3.90] \\
& Consistency & 10 & 3.30 & 0.82 & 3.5 & [2.71, 3.89] \\
& Completeness & 10 & 3.00 & 1.16 & 3.5 & [2.17, 3.83] \\
\midrule
\multirow{4}{*}{Real Data} 
& Linguistic Quality & 10 & 1.80 & 1.14 & 1.0 & [0.99, 2.61] \\
& Information Richness & 10 & 1.90 & 1.10 & 1.5 & [1.11, 2.69] \\
& Consistency & 10 & 3.30 & 0.95 & 4.0 & [2.62, 3.98] \\
& Completeness & 10 & 2.20 & 0.92 & 2.0 & [1.54, 2.86] \\
\bottomrule
\end{tabular}
\vspace{-0.5em}
\caption{\textbf{Descriptive Statistics of Expert Likert Ratings.} Patient-Zero outperforms Real Data across multiple dimensions.}
\label{tab:app_chi_square}
\end{table*}

\begin{table*}[t]
\centering
\small
\begin{tabular}{lccccc}
\toprule
\textbf{Source} &
\textbf{Linguistic} &
\textbf{Diversity} &
\textbf{Consistency} &
\textbf{Completeness} &
\textbf{Overall} \\
\midrule
Real Data &
$2.95\pm1.54$ &
$3.45\pm1.15$ &
$3.70\pm0.98$ &
$3.15\pm1.27$ &
$3.31\pm1.25$ \\

MERA-Qwen &
$\mathbf{3.55\pm1.10}$ &
$3.50\pm1.00$ &
$3.55\pm1.43$ &
$3.40\pm0.94$ &
$3.50\pm1.11$ \\

Patient-Zero &
$3.50\pm0.76$ &
$\mathbf{3.85\pm0.75}$ &
$\mathbf{3.90\pm0.85}$ &
$\mathbf{4.00\pm0.65}$ &
$\mathbf{3.81\pm0.76}$ \\
\bottomrule
\end{tabular}
\vspace{-0.5em}
\caption{\textbf{Physician Evaluation Results.} Values are reported as mean $\pm$ standard deviation.}
\label{tab:app_human_stats}
\end{table*}

\subsection{Evaluation Protocol}
We recruited four senior licensed physicians from top-tier hospitals to conduct a blinded evaluation. The evaluation consisted of two distinct tasks:

\begin{itemize}[left=0pt]
    \item \textbf{Task 1: Human-AI Data Discrimination Test.} Experts were presented with a total of 112 blinded judgments and were asked to classify the origin of each record as either ``Human-authored'' or ``AI-generated''.
    
    \item \textbf{Task 2: Multi-dimensional Quality Assessment.} Experts rated 30 distinct records on a 5-point Likert scale across four clinical dimensions: \textbf{1) Linguistic Quality:} Fluency, professional tone, and absence of redundancy. \textbf{2) Diversity (Information Richness):} Depth of medical details (e.g., specific symptom attributes, history). \textbf{3) Consistency:} Logical coherence between symptoms, diagnosis, and treatment.\textbf{4) Completeness:} Presence of all necessary clinical sections (e.g., Chief Complaint, HPI, Physical Exam).
\end{itemize}

\subsection{Statistical Analysis of Expert Ratings}
We performed pairwise Mann-Whitney U tests to quantify the differences between data sources. Table~\ref{tab:app_human_stats} details the descriptive statistics and significance tests.

\paragraph{Performance against Real Data}
Real Data received the lowest scores in \textit{Linguistic Quality} (Mean=1.80, SD=1.14) and \textit{Completeness} (Mean=2.20, SD=0.92), reflecting the inherent noise in real-world clinical notes (e.g., abbreviations, typos, missing sections). 
In contrast, \textbf{Patient-Zero} achieved significantly higher scores than Real Data in \textit{Information Richness} ($U=85.5, p=0.006, \text{Cohen's } d=1.63$) and comparable scores in other metrics. 
This statistical evidence confirms that our synthetic data not only matches but in specific aspects surpasses the quality of raw human-authored records.

\paragraph{Rater Reliability Analysis}
To ensure the objectivity of the human evaluation, we examined the scoring distribution differences between the annotators. A \textit{Kruskal-Wallis test} revealed no statistically significant difference in their rating patterns ($H=0.054, p=0.8167$). The high p-value indicates strong inter-rater consistency, confirming that the observed performance gaps are attributable to model quality rather than annotator bias.

\subsection{Qualitative Case Studies}
To understand the specific strengths and weaknesses of each source, we analyzed the qualitative feedback provided by the physicians.

\paragraph{Critique of Real Data} Physicians frequently noted that real records were chaotic or incomplete. For example, one evaluator commented on a real record: \textit{``Treatment plan is very chaotic... stopped drug for 3 days but didn't say which drug.''} Another noted: \textit{``Incomplete examination results.''} These comments validate our premise that raw real-world data often lacks the structural integrity required for high-quality model training.

\paragraph{Critique of Baselines (MERA-Qwen)} The baseline model suffered from logical repetition and translation artifacts. Evaluators noted: \textit{``Past history has repetition''} and \textit{``Diagnosis has Broadwater prefix, likely a translation error.''}

\paragraph{Patient-Zero Performance} Our framework produced the most idealized records. Physicians explicitly evaluated the generated cases as \textit{``overall complete and accurate''}, noting that critiques were mostly limited to minor omissions rather than logical contradictions. For instance, one comment pointed out a specific missing detail: \textit{``No description of fever in the disease course.''} This contrast suggests that \textbf{Patient-Zero} successfully captures the high-level structure and logic of clinical narratives, ensuring holistic validity with only fine-grained details occasionally requiring refinement.

\paragraph{Representative Physician-Identified Error}
A representative error was observed in a Patient-Zero bacterial meningitis record.
Fever was included in the structured symptom list but was not explicitly integrated into the longitudinal illness-course narrative.
Physicians therefore downgraded the record for both completeness and consistency.
We categorize this error as a \textit{symptom-to-trajectory anchoring failure}: a record can remain globally clinically plausible while omitting a locally important cross-section dependency.
Based on this error analysis, we introduce explicit cross-section checks requiring key symptoms in the structured state to be reflected consistently in the longitudinal narrative.

% \begin{table}[h]
% \centering
% \caption{Representative Expert Comments on Clinical Records.}
% \label{tab:app_comments}
% \begin{tabular}{ll}
% \toprule
% \textbf{Source} & \textbf{Physician Comment} \\
% \midrule
% \multirow{3}{*}{Real Data} & ``Treatment plan is very chaotic..., stopped drug for 3 days?'' \\
% & ``Incomplete examination results.'' \\
% & ``Logic of Chief Complaint is incorrect.'' \\
% \midrule
% \multirow{2}{*}{MERA-Qwen} & ``Past history has repetition.'' \\
% & ``Diagnosis section has translation artifacts (Broadwater).'' \\
% \midrule
% \multirow{2}{*}{\textbf{Patient-Zero}} & ``Missing description of fever in disease course.'' \\
% & ``Overall complete and accurate.'' \\
% \bottomrule
% \end{tabular}
% \end{table}

\subsection{Source Provenance and Authority}
\label{app:source-provenance}

Patient-Zero does not use real patient records as generation seeds. Instead, the generation pipeline is grounded in curated medical knowledge sources and epidemiological priors. We categorize all external resources into three groups according to their roles in the framework.

\paragraph{Medical knowledge bases}
This category includes Baidu Health Encyclopedia~\cite{baidu_health}, and Chinese Medical Ace resources~\cite{CMA-Base}. These sources are used to construct disease--symptom logic, contraindications, severity criteria, diagnostic criteria, and treatment-related clinical constraints. We select these resources because their entries are authored, curated, or reviewed by licensed physicians, clinical experts, or tertiary-hospital medical teams. In Patient-Zero, they provide the clinical logic required for disease outline standardization and symptom progression. Overall, this category supports the disease set and standardized outlines used in our 98-disease, six-specialty generation corpus.

% \paragraph{Peer-reviewed medical literature}
% This category includes peer-reviewed journals and authoritative clinical or epidemiological sources, such as NEJM, The Lancet, Nature Medicine, specialty guidelines, and population-level epidemiological surveys where available. These sources are used to derive epidemiological priors, demographic and lifestyle distributions, disease-specific prevalence, and clinically relevant risk relations. We select them because they are published in peer-reviewed venues, issued by professional societies, or based on authoritative population-level surveys. In Patient-Zero, they provide the statistical priors and risk dependencies used for epidemiological attribute permutation. Overall, this category contains XX references, covers XX clinical specialties, and supports XX epidemiological priors.

\paragraph{Public structured references.}
This category includes MIMIC-IV, Synthea, SCRIPT, CMA, and other public benchmark datasets. These resources are used only as baselines or comparison references in evaluation, and are never used as generation seeds. We select them because they are publicly documented benchmark datasets with established use in clinical NLP, EHR modeling, or synthetic-data evaluation. In Patient-Zero, they support baseline comparison, distributional analysis, and evaluation calibration, but do not contribute patient-level seed records to the generation pipeline.

\section{Case Studies and Prompts}
\label{app:case_study}

\begin{table*}[t]
\centering
\begin{tcolorbox}[
  colback=teal!5,
  colframe=teal!60!black,
  title=\textbf{[Synthetic Case 1] Judged as Human-authored by Experts: Neurofibromatosis Type 1},
  fonttitle=\bfseries,
  boxrule=0.6pt,
  arc=1.5mm,
  toptitle=1mm
]
    \small
    % --- Patient Profile ---
    \textbf{Patient Profile}
    \begin{itemize}[nosep]
        \item \textbf{Name:} Claire Thompson
        \item \textbf{Demographics:} 54-year-old Female, Caucasian
        \item \textbf{Socioeconomic:} Suburban residence, Low Income
        \item \textbf{Record ID:} 398899a3-8d31-4805-8279-4361209810d9
    \end{itemize}
    
    \tcblower 
    \small
    % --- Presenting Complaint ---
    \textbf{Presenting Complaint: Cutaneous Lesions \& Spinal Pain}
    \begin{itemize}[nosep]
         \item \textbf{Symptoms:}  Café-au-lait spots (since childhood), Axillary freckling, Cutaneous neurofibromas  (slow increase since 20s), Dome-shaped soft nodules, Intermittent spinal pain.
         \item \textbf{History:} Lesions largely stable; mild surge during pregnancies. Spinal pain worse with sitting, relieved by stretching.
         \item \textbf{Family History:} Father with NF1; Mother with Hypertension.
    \end{itemize}

    {\color{black!20}\rule{\linewidth}{0.5pt}} 
    
    % --- Medical History ---
    \textbf{Medical History}
    \begin{itemize}[nosep]
        \item \textbf{Diagnosis:} NF1 diagnosed at age 28 ($\geq$6 café-au-lait macules + freckling).
        \item \textbf{Comorbidities:} Mild thoracic scoliosis 

[Image of thoracic scoliosis X-ray]
 (adolescence), Hypertension risk (monitored).
        \item \textbf{Interventions:} Conservative dermatology follow-up; no prior surgeries.
    \end{itemize}

    {\color{black!20}\rule{\linewidth}{0.5pt}} 

    % --- Examination Results ---
    \textbf{Examination Results}
    \begin{itemize}[nosep]
        \item \textbf{Genetics:} Targeted next-generation sequencing of NF1/NF2 (mean on-target coverage 312x; 99.2\% of coding bases >50x) identified a heterozygous truncating variant in NF1: c.2041C>T (p.Arg681*), variant allele fraction 47\% in peripheral blood leukocytes. No pathogenic or likely pathogenic variants were detected in NF2. Copy number analysis showed no exon-level deletions/duplications in NF1 or NF2. Sanger confirmation of the NF1 variant was positive. Microarray (SNP/CGH) was unremarkable.
        \item \textbf{Ultrasound:} High-frequency (15 MHz) soft tissue ultrasound of trunk and upper arms: 14 discrete subcutaneous nodules, ovoid, well-circumscribed, hypoechoic relative to surrounding fat (mean echogenicity 19–23 grayscale units vs fat ~34–38), located within dermis/subcutis at depths 2–6 mm. Largest lesion: right posterior shoulder 1.2 x 0.9 x 0.7 cm; majority 0.4–0.8 cm. Lesions are compressible and mobile over deep fascia; no deep fascial involvement. Color Doppler shows minimal internal vascularity in 3/14 lesions (scattered low-flow signals; peak systolic velocity 4–6 cm/s; resistive index 0.56–0.62). Axillary region: skin freckling noted externally; no discrete mass on sonography. Focused abdominal and pelvic survey: liver (size 15.8 cm, homogeneous), spleen (10.9 cm), kidneys (R 10.8 cm, L 10.6 cm), uterus and adnexa within normal sonographic limits; no intra-abdominal solid masses. 
        \item \textbf{MRI Examination:} Thoracic spine MRI (1.5T): At right T9 dorsal root region, an ovoid lesion measuring 7 x 6 x 5 mm along the nerve sheath. T1-weighted images: isointense to skeletal muscle (signal intensity ratio lesion/muscle ~1.05). T2-weighted images: markedly hyperintense with a central slightly low-signal area (target sign) occupying ~42\% of the cross-sectional area. STIR: strong hyperintensity. Post-gadolinium T1: mild heterogeneous enhancement (signal increase ~38\%). No spinal cord edema, no mass effect; canal remains patent. Additional findings: mild degenerative changes with L4–L5 posterior disc protrusion 2 mm, no nerve root contact; thoracic coronal localizer shows mild scoliosis measuring 11°.
        \item \textbf{Spinal Angiography:} Selective catheter spinal angiography: right T9 intercostal/radiculomedullary artery injection demonstrates normal arterial phase and opacification of the anterior spinal artery without arteriovenous shunting. A subtle contrast column indentation is observed at the right nerve root exit zone, consistent with a small extrinsic filling defect. No abnormal tumor blush; contrast remains intrathecal without extravasation. No accidental contrast injection into a mass. Other levels show normal vascular anatomy.
         \item \textbf{Ophthalmic Examination:} Best-corrected visual acuity: OD 20/25, OS 20/30 (with +1.50 D add for near). Intraocular pressure: OD 14 mmHg, OS 13 mmHg (Goldmann applanation). 
         Slit-lamp: multiple raised, smooth, brown-yellow iris nodules consistent with hamartomas-OD: 6 nodules ranging 0.8-1.5 mm; OS: 5 nodules ranging 0.7-1.3 mm; located predominantly in inferior and temporal quadrants. Cornea and anterior chamber clear. 
         Optic discs sharp with cup-to-disc ratio 0.3 OU; no edema, pallor, or atrophy. Optical coherence tomography RNFL: average thickness OD 99 $\mu$m, OS 97 $\mu$m; macular cube thickness OD 279 $\mu$m, OS 276 $\mu$m. 
         Humphrey 24-2 visual fields: MD OD -0.8 dB, OS -1.1 dB; pattern deviation without focal defects.
         \item \textbf{X-ray Examination:} Standing PA and lateral thoracolumbar radiographs: mild right-convex thoracic scoliosis from T6–T11 with Cobb angle 12°. Vertebral body heights maintained; pedicles symmetric; no ribbon-like lucencies or longitudinal stripe-like lesions within bones. Mild multilevel spondylotic changes (small anterior osteophytes at T8–T10 and L4–L5). No vertebral scalloping, no dysplasia of long bones, ribs, or pelvis. Overall skeletal involvement minimal on plain films.
    \end{itemize}
\end{tcolorbox}
\vspace{-10pt}
\caption{\textbf{Synthetic Patient Record (Neurofibromatosis Type 1).} This record was judged as \textbf{\textit{human-authored"}} by the licensed physician. It presents a classic, longitudinal clinical picture of Neurofibromatosis Type 1 in a middle-aged female, capturing the typical evolution of cutaneous markers (pregnancy-associated progression), specific ophthalmological criteria (Lisch nodules), and characteristic radiological findings (Target sign on MRI).}
\label{fig:claire_thompson_case}
\end{table*}

\begin{table*}[t]
\centering
\begin{tcolorbox}[
  colback=teal!5,
  colframe=teal!60!black,
  title=\textbf{[Synthetic Case 2] Judged as Human-authored by Experts: Pneumonia},
  fonttitle=\bfseries,
  boxrule=0.6pt,
  arc=1.5mm,
  toptitle=1mm
]
    \small
    % --- Patient Profile ---
    \textbf{Patient Profile}
    \begin{itemize}[nosep]
        \item \textbf{Name:} Rohan Kumar
        \item \textbf{Demographics:} 21-year-old Male, Asian
        \item \textbf{Socioeconomic:} Rural residence, Middle Income
        \item \textbf{Record ID:} ef536bce-a379-4a50-9140-d889740195bd
    \end{itemize}
    
    \tcblower 
    \small
    
    % --- Presenting Complaint ---
    \textbf{Presenting Complaint: Respiratory Infection (Mild Severity)}
    \begin{itemize}[nosep]
         \item \textbf{Symptoms:} Productive cough (yellow purulent sputum), low-grade fever, mild pleuritic chest pain, fatigue, loss of appetite.
         \item \textbf{History:} Cold-like onset 5 days ago; progressed to productive cough. Stable/improving with OTC care.
         \item \textbf{Risk Factors:} Current smoker (7 cigs/day), Indoor wood-smoke exposure.
    \end{itemize}

    {\color{black!20}\rule{\linewidth}{0.5pt}} 
    
    % --- Medical History ---
    \textbf{Medical History}
    \begin{itemize}[nosep]
        \item \textbf{Comorbidities:} Seasonal allergic rhinitis. History of acute bronchitis (age 19).
        \item \textbf{Lifestyle:} Vegetarian diet. Heavy alcohol use (weekend binge).
    \end{itemize}

    {\color{black!20}\rule{\linewidth}{0.5pt}} 

    % --- Examination Results ---
    \textbf{Examination Results}
    \begin{itemize}[nosep]
        \item \textbf{Complete Blood Count:} Collected at 08:20 on day 5 of illness. WBC $11.6 x10^9/L$ (ref 4.0–10.0), with neutrophils 78\% (absolute neutrophil count $9.0 x10^9/L$), lymphocytes 15\% $(1.7 x10^9/L)$, monocytes 5\% $(0.58 x10^9/L)$, eosinophils 1\% $(0.12 x10^9/L)$, basophils 1\% $(0.12 x10^9/L)$. Bands 3\% noted. Hemoglobin 14.3 g/dL, hematocrit 42.1\%, MCV 87 fL, MCH 29 pg, RDW 12.9\%. Platelets $325 x10^9/L$. Findings indicate mild leukocytosis with neutrophil predominance.
        \item \textbf{C-Reactive Protein and Procalcitonin:} Serum CRP 38 mg/L (ref <5 mg/L). Procalcitonin 0.12 ng/mL (ref <0.05 ng/mL; values <0.25 ng/mL generally associated with milder bacterial activity). Repeat CRP at 48 hours planned; current trend reported by patient (symptomatic) is slightly improving with antipyretics.
        \item \textbf{Oxygenation Assessment and Arterial Blood Gas Analysis:} Vital signs at rest (room air, FiO2 0.21): Temp 37.9°C, HR 88 bpm, RR 18/min, BP 118/72 mmHg. Pulse oximetry SpO2 97–98\% at rest; with brisk 3-minute walk SpO2 95–96\% and RR 20/min, recovery to 97\% within 1 minute. Arterial blood gas (right radial artery, room air): pH 7.44, PaCO2 36 mmHg, PaO2 92 mmHg, HCO3- 24 mmol/L, Base excess +0.5 mmol/L, SaO2 97\%. Calculated alveolar-arterial (A–a) O2 gradient ~13 mmHg (slightly above age-adjusted expected). Lactate 1.2 mmol/L (ref 0.5–2.2). Carboxyhemoglobin 2.1\% (consistent with current smoker), Methemoglobin 0.5\%. Overall oxygenation adequate without resting hypoxemia.
        \item \textbf{Sputum Smear, Sputum Culture, and Drug Sensitivity Test:} Early-morning expectorated sputum, 3 mL, yellow mucopurulent. Gram stain: >25 polymorphonuclear leukocytes/LPF, <10 squamous epithelial cells/LPF (good-quality specimen). Predominant gram-positive lancet-shaped diplococci observed; occasional gram-negative rods; no acid-fast bacilli; fungal elements not seen. Culture (48 hours): Moderate growth (2+) of alpha-hemolytic colonies on 5\% sheep blood agar; optochin zone 20 mm and bile solubility positive, consistent with Streptococcus pneumoniae. Normal respiratory flora also present; no growth of Staphylococcus aureus or Pseudomonas aeruginosa. Antimicrobial susceptibility (MIC, CLSI categories): Penicillin (parenteral) 0.06 mg/L – S; Amoxicillin/clavulanate 0.5 mg/L – S; Ceftriaxone 0.25 mg/L – S; Levofloxacin 1 mg/L – S; Erythromycin 8 mg/L – R; Azithromycin 8 mg/L – R; Doxycycline 2 mg/L – I; Trimethoprim-sulfamethoxazole >4 mg/L – R; Vancomycin 0.5 mg/L – S; Linezolid 1 mg/L – S.
        \item \textbf{Imaging Examination:} Chest X-ray (PA and lateral) performed on day 5 of illness. Cardiomediastinal silhouette normal; cardiothoracic ratio 0.49. Lungs: patchy air-space opacity in the right lower lobe, posterior basal segment, measuring approximately 3.2 x 2.1 cm, with faint air bronchograms and mild peribronchial thickening in the right infrahilar region. No pleural effusion; costophrenic angles sharp. No cavitation or pneumothorax. Lateral view localizes opacity to the posterior segment of the right lower lobe. Findings consistent with a focal infectious/inflammatory process of limited extent.
    \end{itemize}
\end{tcolorbox}
\vspace{-10pt}
\caption{\textbf{Synthetic Patient Record (Pulmonology).} This record was judged as judged as\textbf{\textit{ human-authored}} by licensed physician. It accurately simulates a typical mild Community-Acquired Pneumonia (CAP) in a young adult, featuring the classic progression from viral prodrome to bacterial superinfection, coherent microbiological findings (pneumococcal diplococci), and consistent radiographic evidence (lobar consolidation).}
\label{fig:rohan_kumar_case}
\end{table*}

\begin{table*}[t]
\centering
\begin{tcolorbox}[
  colback=yellow!5,
  colframe=yellow!60!black,
  title=\textbf{Rejected Sample Analysis: 76-year-old Male of Colorectal Cancer},
  fonttitle=\bfseries,
  boxrule=0.6pt,
  arc=1.5mm,
  toptitle=1mm
]
\small
    % --- Patient Profile ---
    \textbf{Patient Profile}
    \begin{itemize}[nosep]
        \item \textbf{Name:} Arthur Whitaker
        \item \textbf{Demographics:} 76-year-old Male, Caucasian
        \item \textbf{Socioeconomic:} Rural residence, Low Income
        \item \textbf{Target Severity Label:} \textbf{Mild} % 强调冲突点
    \end{itemize}
    
    \tcblower 
    \small
    % --- Synthesized Content ---
    \textbf{Synthesized Clinical Narrative (Draft)}
    \begin{itemize}[nosep]
         \item \textbf{Symptoms:} Change in bowel habits, bloody stool, abdominal pain, diarrhea. Notably reports \textbf{progressive fatigue}, \textbf{unintentional weight loss (3kg)}, and \textbf{loss of appetite}.
         \item \textbf{History:} Symptoms began 3–4 months ago. Initial rectal bleeding followed by constitutional symptoms. Lab work revealed microcytic iron-deficiency anemia.
         \item \textbf{Diagnosis:} Colonoscopy confirmed localized left-sided adenocarcinoma.
    \end{itemize}

    {\color{black!20}\rule{\linewidth}{0.5pt}} 
    
    % --- Rejection Logic ---
    \textbf{Rejection Logic}
    \begin{itemize}[nosep]
        \item \textbf{Conflict Type:} Severity Mismatch.
        \item \textbf{Reasoning:} The generated narrative includes significant constitutional symptoms (\textit{weight loss, anorexia, fatigue}) and systemic complications (\textit{iron-deficiency anemia}). These clinical features indicate a disease burden that exceeds the clinical definition of ``Mild'' for early-stage colorectal cancer, which is typically asymptomatic or presents with minor local symptoms only.
        \item \textbf{Outcome:} \textbf{\textit{Rejected}} by NLI-Verifier. Triggered regeneration with strict constraints on systemic symptoms.
    \end{itemize}
\end{tcolorbox}
\vspace{-10pt}
\caption{\textbf{Quality Control Visualization.} An example of a synthetic record rejected by our framework. The \textit{NLI-Verifier} detected a semantic contradiction where the generated symptom burden (systemic constitutional symptoms) conflicted with the input condition of ``Mild'' severity, ensuring strict clinical consistency in the final dataset.}
\label{fig:rejected_case_crc}
\end{table*}

\begin{table*}[t]
\centering
\begin{tcolorbox}[
  colback=yellow!5,
  colframe=yellow!60!black,
  title=\textbf{Rejected Sample Analysis: 44-year-old Female of Cirrhosis},
  fonttitle=\bfseries,
  boxrule=0.6pt,
  arc=1.5mm,
  toptitle=1mm
]
    \small
    % --- Patient Profile ---
    \textbf{Patient Profile}
    \begin{itemize}[nosep]
        \item \textbf{Name:} Layla Al-Khatib
        \item \textbf{Demographics:} 44-year-old Female, Middle Eastern
        \item \textbf{Socioeconomic:} Suburban residence, Low Income
        \item \textbf{Target Condition:} \textbf{Compensated Cirrhosis (Mild)}
    \end{itemize}
    
    \tcblower 
    \small
    % --- Synthesized Content ---
    \textbf{Synthesized Clinical Narrative (Draft)}
    \begin{itemize}[nosep]
         \item \textbf{Medical History:} History of alcohol use disorder. Compensated cirrhosis diagnosed 1 year ago based on ultrasound showing coarse liver and mild splenomegaly \textbf{without ascites}.
         \item \textbf{Generated Symptoms:} Fatigue, loss of appetite, weight loss, and \textbf{abdominal distension}.
         \item \textbf{Narrative Detail:} "She notes mild \textbf{abdominal fullness/distension} for the past 6 months... No episodes of jaundice or ascites to date."
    \end{itemize}

    {\color{black!20}\rule{\linewidth}{0.5pt}} 
    
    % --- Rejection Logic ---
    \textbf{Rejection Logic}
    \begin{itemize}[nosep]
        \item \textbf{Conflict Type:} Logical Contradiction (Pathophysiology).
        \item \textbf{Reasoning:} The generated narrative lists "abdominal distension" as a symptom while explicitly stating the patient has "no ascites" and is in the "compensated" stage. In cirrhosis, abdominal distension is primarily caused by fluid accumulation (ascites). Asserting distension in the absence of ascites (or gas/obesity, which were not contextualized) creates a medically incoherent presentation for a liver disease case.
        \item \textbf{Outcome:} \textbf{\textit{Rejected}} by NLI-Verifier. The record was flagged for factual inconsistency regarding signs of decompensation.
    \end{itemize}
\end{tcolorbox}
\vspace{-10pt}
\caption{\textbf{Logic Consistency Check.} In this rejected sample, our framework detected a conflict between the generated symptom (\textit{abdominal distension}) and the clinical constraints (\textit{no ascites, compensated stage}), preventing the generation of physiologically impossible patient records.}
\label{fig:rejected_case_cirrhosis}
\end{table*}

\begin{table*}[t]
\centering
\begin{tcolorbox}[
  colback=gray!5,                
  colframe=gray!60!black,        
  title=\textbf{Prompt Template: Medically-Aligned Disease Outline Generation},
  fonttitle=\bfseries,
  boxrule=0.6pt,
  arc=1.5mm,
  toptitle=1mm,
  bottomtitle=1mm
]
    \small
    % --- System Instruction ---
    \textbf{System Instruction}
    \begin{itemize}[nosep, leftmargin=1em]
        \item \textbf{Role:} Senior Medical Expert.
        \item \textbf{Task:} Analyze raw disease info and generate a refined JSON outline for synthetic record generation.
        \item \textbf{Output Format:} \\
        STRICT JSON format containing: \texttt{disease\_summary}, \texttt{key\_characteristics}, \texttt{typical\_presentation} (mild/moderate/severe), \texttt{important\_notes}, \texttt{contraindications}, \texttt{differential\_considerations}, \texttt{special\_populations}, and \texttt{red\_flags}.
    \end{itemize}
    
    \tcblower 
    \small
    % --- User Input ---
    \textbf{User Prompt Template}
    \begin{itemize}[nosep, leftmargin=1em]
         \item \textbf{Context Injection:} Analyze the following information:
         \begin{itemize}[nosep]
             \item \texttt{\{raw\_outline\}}: [Raw medical knowledge text]
             \item \texttt{\{profile\_context\}}: [e.g., Demographics, Risk Factors]
             \item \texttt{\{severity\_context\}}: [e.g., Target Severity Level]
         \end{itemize}
    \end{itemize}

    \vspace{4pt}
    {\color{black!20}\rule{\linewidth}{0.5pt}} 
    \vspace{2pt}
    
    % --- Constraint Checklist ---
    \textbf{Generation Instructions \& Constraints}
    \begin{enumerate}[nosep, leftmargin=1.2em]
        \item Summarize key clinical characteristics concisely.
        \item Describe typical presentations across severity levels (Mild, Moderate, Severe).
        \item List \textbf{Important Notes} for generation: Age/Gender manifestations, symptom timing, comorbidities.
        \item Identify \textbf{Contraindications}: Incompatible demographics, conflicting symptoms, unrealistic findings.
        \item Note considerations for \textbf{Special Populations} (Pediatric, Elderly, Pregnant).
        \item List \textbf{Red Flags} indicating severe complications.
    \end{enumerate}
\end{tcolorbox}
\vspace{-10pt}
\caption{\textbf{Disease Outline Generation Prompt.} The prompt guides the LLM to structure unstructured medical knowledge into a hierarchical JSON format, enforcing clinical constraints and severity-specific presentations before the actual patient record generation begins.}
\label{tab:prompt_outline}
\end{table*}

\begin{table*}[t]
\centering
\begin{tcolorbox}[
  colback=gray!5,                 % Background: Light Gray
  colframe=gray!60!black,         % Frame: Dark Gray
  title=\textbf{Prompt Template: Patient Record \& Symptom Generation},
  fonttitle=\bfseries,
  boxrule=0.6pt,
  arc=1.5mm,
  toptitle=1mm,
  bottomtitle=1mm
]
    \small
    % --- System Instruction ---
    \textbf{System Instruction}
    \begin{itemize}[nosep, leftmargin=1em]
        \item \textbf{Role:} Knowledgeable Medical Expert.
        \item \textbf{Task:} Generate a comprehensive patient record with realistic symptoms based on the provided profile.
    \end{itemize}

    \vspace{4pt}
    \textbf{Input Data Injection}
    \begin{itemize}[nosep, leftmargin=1em]
         \item \textbf{Target Condition:} \texttt{\{disease\_name\}}, \texttt{\{severity\_level\}}
         \item \textbf{Demographic Profile:} 
         \begin{itemize}[nosep]
            \item Biological: \texttt{\{Age\}}, \texttt{\{Sex\}}, \texttt{\{Physiological\_Status\}}
            \item Sociocultural: \texttt{\{Ethnicity\}}, \texttt{\{Geography\}}, \texttt{\{Socioeconomic\}}
            \item Behavioral: \texttt{\{Smoking\}}, \texttt{\{Alcohol\}}, \texttt{\{Diet\}}, \texttt{\{Lifestyle\}}
         \end{itemize}
         \item \textbf{Clinical Knowledge Base:}
         \begin{itemize}[nosep]
            \item \texttt{\{symptoms\_list\}}: List of possible symptoms for this disease.
            \item \texttt{\{disease\_outline\}}: Structured characteristics and progression logic.
         \end{itemize}
    \end{itemize}
    
    \tcblower 
    \small
    % --- Generation Directives ---
    \textbf{Generation Directives}
    \begin{enumerate}[nosep, leftmargin=1.2em]
        \item \textbf{Identity Synthesis:} Generate a realistic name reflecting the patient's ethnicity and country of origin.
        \item \textbf{History Construction:} Create a detailed medical history consistent with the disease and age profile.
        \item \textbf{Background Factors:} Synthesize lifestyle factors, vaccination history, and family history.
        \item \textbf{Symptom Selection:} Select specific symptoms from the provided list appropriate for the target \texttt{severity}.
        \item \textbf{Chronology:} Specify symptom duration and progression logic.
    \end{enumerate}

    {\color{black!20}\rule{\linewidth}{0.5pt}}

    \textbf{Output Requirement:} Output ONLY a valid JSON object.
\end{tcolorbox}
\vspace{-10pt}
\caption{\textbf{Patient Record Generation Prompt.} This core module synthesizes the abstract demographic parameters (e.g., "Middle Income", "Smoker") and medical knowledge into a concrete patient identity (e.g., "Name", "Specific History"), ensuring the resulting profile is culturally consistent and clinically plausible.}
\label{tab:prompt_patient_gen}
\end{table*}

\begin{table*}[t]
\centering
\begin{tcolorbox}[
  colback=gray!5,                 % Background: Light Gray
  colframe=gray!60!black,         % Frame: Dark Gray
  title=\textbf{Prompt Template: Examination Result Generation},
  fonttitle=\bfseries,
  boxrule=0.6pt,
  arc=1.5mm,
  toptitle=1mm,
  bottomtitle=1mm
]
    \small
    % --- System Instruction ---
    \textbf{System Instruction}
    \begin{itemize}[nosep, leftmargin=1em]
        \item \textbf{Role:} Knowledgeable Medical Expert.
        \item \textbf{Task:} Generate specific, quantitative examination findings based on patient symptoms and disease severity.
    \end{itemize}

    \vspace{4pt}
    \textbf{Input Data Injection}
    \begin{itemize}[nosep, leftmargin=1em]
         \item \textbf{Case Context:} 
         \begin{itemize}[nosep]
            \item Patient: \texttt{\{patient\_info\}}, \texttt{\{symptoms\_data\}}
            \item Disease: \texttt{\{disease\_name\}}, \texttt{\{severity\_level\}}
         \end{itemize}
         \item \textbf{Medical Grounding:}
         \begin{itemize}[nosep]
            \item \texttt{\{exam\_list\}}: List of required examinations (e.g., CBC, CT Scan).
            \item \texttt{\{exam\_reference\}}: \textbf{Reference standards} and normal ranges for each exam.
            \item \texttt{\{disease\_outline\}}: Pathophysiological characteristics guide.
         \end{itemize}
    \end{itemize}
    
    \tcblower 
    \small
    % --- Generation Directives ---
    \textbf{Generation Directives}
    \begin{enumerate}[nosep, leftmargin=1.2em]
        \item \textbf{Quantitative Specificity:} Include realistic values, percentages, and measurements (e.g., ``Hemoglobin 8.5 g/dL'' rather than ``Low Hemoglobin'').
        \item \textbf{Severity Alignment:} Findings must strictly reflect the \texttt{\{severity\}} (e.g., Mild fibrosis vs. Decompensated ascites).
        \item \textbf{Diagnosis Blinding:} \textbf{Do NOT} directly mention the disease name in the results. Describe the \textit{signs}, not the \textit{label}.
        \item \textbf{Clinical Realism:} Mimic the observation style of real clinical reports.
    \end{enumerate}

    {\color{black!20}\rule{\linewidth}{0.5pt}} 

    % --- Output Format ---
    \textbf{Output Requirement:} Output ONLY a valid JSON object.

\end{tcolorbox}
\vspace{-10pt}
\caption{\textbf{Examination Generation Prompt.} Unlike open-ended generation, this module is grounded by injected \texttt{Reference Standards}, forcing the LLM to generate precise numerical data and observations that are statistically plausible within the specific clinical context.}
\label{tab:prompt_exam_gen}
\end{table*}

\begin{table*}[t]
\centering
\begin{tcolorbox}[
  colback=gray!5,                 % 背景色：浅灰
  colframe=gray!60!black,         % 边框色：深灰
  title=\textbf{Prompt Template: Clinical Symptom and Consistency Validation},
  fonttitle=\bfseries,
  boxrule=0.6pt,
  arc=1.5mm,
  toptitle=1mm,
  bottomtitle=1mm
]
    \small
    % --- System Instruction & Context ---
    \textbf{System Instruction}
    \begin{itemize}[nosep, leftmargin=1em]
        \item \textbf{Role:} Medical Expert Reviewer.
        \item \textbf{Task:} Review a generated patient record for clinical accuracy and guideline compliance.
    \end{itemize}

    \vspace{4pt}
    \textbf{Input Data Injection}
    \begin{itemize}[nosep, leftmargin=1em]
         \item \textbf{Disease Metadata:} \texttt{\{disease\_name\}}, \texttt{\{severity\_level\}}
         \item \textbf{Patient Profile:} 
         \begin{itemize}[nosep]
            \item Age: \texttt{\{age\}} (\texttt{\{age\_group\}})
            \item Sex: \texttt{\{sex\}}
            \item Physiological Status: \texttt{\{physiological\_status\}}
            \item Smoking Status: \texttt{\{smoking\_status\}}
         \end{itemize}
         \item \textbf{Generated Record:} \texttt{\{json.dumps(response)\}}
    \end{itemize}
    
    \tcblower 
    \small
    % --- Validation Logic ---
    \textbf{Validation Criteria (Medical Knowledge)}
    \begin{enumerate}[nosep, leftmargin=1.2em]
        \item \textbf{Symptom Consistency:} Symptoms must be clinically consistent with the disease and severity.
        \item \textbf{Severity Matching:} Symptom count/intensity must match the stated level (Mild vs. Severe).
        \item \textbf{Demographic Logic:} Age/Sex must be logically consistent with medical history.
        \item \textbf{Contradiction Check:} No impossible combinations (e.g., pregnant male, pediatric conditions in elderly).
        \item \textbf{Realistic Progression:} Duration description should be realistic for the disease.
        \item \textbf{Plausible History:} Medical history must align with age and condition.
    \end{enumerate}

    {\color{black!20}\rule{\linewidth}{0.5pt}} 

    % --- Output Format ---
    \textbf{Output Requirement}
    \begin{itemize}[nosep, leftmargin=1em]
        \item Evaluate strict compliance with the above knowledge.
        \item \textbf{Format:} Output ONLY a JSON object.
    \end{itemize}
\end{tcolorbox}
\vspace{-10pt}
\caption{\textbf{Symptom Validation Prompt.} This prompt acts as an automated critic, employing a frozen LLM to verify that the generated patient narrative strictly adheres to the input constraints and clinical pathophysiology before the record is accepted into the dataset.}
\label{tab:prompt_validation}
\end{table*}

\begin{table*}[t]
\centering
\begin{tcolorbox}[
  colback=gray!5,                 % Background: Light Gray
  colframe=gray!60!black,         % Frame: Dark Gray
  title=\textbf{Prompt Template: Examination Results Verification},
  fonttitle=\bfseries,
  boxrule=0.6pt,
  arc=1.5mm,
  toptitle=1mm,
  bottomtitle=1mm
]
    \small
    % --- System Instruction & Context ---
    \textbf{System Instruction}
    \begin{itemize}[nosep, leftmargin=1em]
        \item \textbf{Role:} Medical Expert Reviewer.
        \item \textbf{Task:} Review generated examination results for clinical accuracy, realism, and knowledge compliance.
    \end{itemize}

    \vspace{4pt}
    \textbf{Input Data Injection}
    \begin{itemize}[nosep, leftmargin=1em]
         \item \textbf{Clinical Context:} \texttt{\{disease\_name\}}, \texttt{\{severity\_level\}}
         \item \textbf{Patient Data:} \texttt{\{patient\_info\}}, \texttt{\{symptoms\_data\}}
         \item \textbf{Target:} 
         \begin{itemize}[nosep]
            \item Generated Results: \texttt{\{exam\_results\}}
            \item Expected Standard: \texttt{\{expected\_exams\}}
         \end{itemize}
    \end{itemize}
    
    \tcblower 
    \small
    % --- Validation Logic ---
    \textbf{Validation Criteria (Medical Knowledge)}
    \begin{enumerate}[nosep, leftmargin=1.2em]
        \item \textbf{Clinical Consistency:} Results must align with the disease pathology and reported symptoms.
        \item \textbf{Realistic Values:} Measurements and observations must be within plausible biological ranges.
        \item \textbf{Severity Reflection:} Results should appropriately reflect the stated severity (e.g., mild vs. severe abnormalities).
        \item \textbf{Findings Only:} Results must NOT directly state the diagnosis (describe the \textit{signs}, not the \textit{conclusion}).
        \item \textbf{Completeness:} All critical examinations for the specific condition must have meaningful entries.
        \item \textbf{Internal Logic:} No contradictory findings between different test modalities (e.g., Lab vs. Imaging).
        \item \textbf{Unit Precision:} Numeric values must use correct medical units.
    \end{enumerate}

    {\color{black!20}\rule{\linewidth}{0.5pt}} 

    % --- Output Format ---
    \textbf{Output Requirement}
    \begin{itemize}[nosep, leftmargin=1em]
        \item Evaluate strict compliance with the above knowledge.
        \item \textbf{Format:} Output ONLY a JSON object.
    \end{itemize}
\end{tcolorbox}
\vspace{-10pt}
\caption{\textbf{Examination Validation Prompt.} This module ensures that generated medical test results (e.g., Labs, Imaging) are not only clinically accurate but also statistically realistic, preventing "diagnosis leakage" where the model prematurely states the conclusion instead of raw findings.}
\label{tab:prompt_exam_validation}
\end{table*}

\end{document}